\documentclass{IEEEoj}
\usepackage{cite}
\usepackage{amsmath,amssymb,amsfonts}
\usepackage{algorithmic}
\usepackage{graphicx,color}
\usepackage{textcomp}
\usepackage{bm}
\usepackage{array}
\usepackage{float}
\usepackage{multicol}
\usepackage{multirow}
\usepackage{makecell}
\usepackage{stfloats}
\usepackage{subfigure}
\usepackage{titlesec}
\let\labelindent\relax
\usepackage{enumitem}
\usepackage{hyperref}
\hypersetup{hidelinks=true}

\titlespacing{\section}{0pt}{5pt}{5pt}
\titlespacing{\subsection}{0pt}{4pt}{4pt}
\titlespacing{\subsubsection}{0pt}{3pt}{3pt}

\newtheorem{proposition}{Proposition}

\newcommand{\tabincell}[2]{\begin{tabular}{@{}#1@{}}#2\end{tabular}}

\def\BibTeX{{\rm B\kern-.05em{\sc i\kern-.025em b}\kern-.08em
    T\kern-.1667em\lower.7ex\hbox{E}\kern-.125emX}}
\AtBeginDocument{\definecolor{tmlcncolor}{cmyk}{0.93,0.59,0.15,0.02}\definecolor{NavyBlue}{RGB}{0,86,125}}
\def\OJlogo{\vspace{-4pt}\includegraphics[height=18pt]{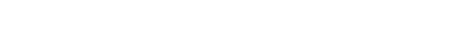}}
\def\seclogo{\vspace{10pt}\includegraphics[height=18pt]{Fig/new_logo}}

\begin{document}

\renewcommand{\thesubsection}{\thesection-\Alph{subsection}}
\renewcommand{\thesubsubsection}{\thesubsection\arabic{subsubsection}}

\title{Designing Heterogeneous GNNs with Desired Permutation Properties for Wireless Resource Allocation}

\author{Jianyu Zhao, Chenyang Yang, Tingting Liu}
\affil{School of Electronics and Information Engineering, Beihang University, Beijing {\rm 100191}, China}
\corresp{CORRESPONDING AUTHOR: Chenyang Yang (e-mail: cyyang@buaa.edu.cn).}

\begin{abstract}
Graph neural networks (GNNs) have been designed for learning a variety of wireless policies, i.e., the mappings from environment parameters to decision variables, thanks to their superior performance, and the potential in enabling scalability and size generalizability. These merits are rooted in leveraging permutation prior, i.e., satisfying the permutation property of the policy to be learned (referred to as desired permutation property).
Many wireless policies are with complicated permutation properties. To satisfy these properties, heterogeneous GNNs (HetGNNs) should be used to learn such policies.
There are two critical factors that enable a HetGNN to satisfy a desired permutation property: constructing
an appropriate heterogeneous graph and judiciously designing the architecture of the HetGNN. However, both the graph and the HetGNN are designed heuristically so far.
In this paper, we strive to provide a systematic approach for the design to satisfy the desired permutation property.
We first propose a method for constructing a graph for a policy, where the edges and their types are defined for the sake of satisfying complicated permutation properties.
Then, we provide and prove three sufficient conditions to design a HetGNN such that it can satisfy the desired permutation property when learning over an appropriate graph.
These conditions suggest a method of designing the HetGNN with desired permutation property by sharing the processing, combining, and pooling functions according to the types of vertices and edges of the graph.
We take power allocation and hybrid precoding policies as examples for demonstrating how to apply the proposed methods and validating the impact of the permutation prior by simulations.
\end{abstract}



\maketitle

\section{Introduction}
Graphs are able to represent complex relations among entities, making them important in many fields such as computer science, social science,  molecular biology and wireless communications.

Graph neural networks (GNNs) have been proposed to learn over graphs \cite{GNNsurvey},
which have witnessed great success in many domains, owing to leveraging topology prior and permutation prior \cite{battaglia2018relational}. The topology prior is the topology information, say the relation among vertices of a graph. This prior is automatically used by a GNN via learning over a graph, since the representation of each vertex or edge is updated only by aggregating the information from its neighboring vertices and edges. The permutation prior is a mathematical property regarding the input-output relation of a multivariate function. If the output of the function permutes accordingly or remains unchanged with the permuted input, then the function exhibits permutation property. A multivariate function defined on sets has permutation property \cite{zaheer2017deep, SEquivariance,maron2019universality,hartford2018deep}. Exploiting the permutation prior indicates that the permutation property induced by all the sets of the function has
been used for designing the GNN.


The role of the two relational priors highly depends on applications. In graph learning domain, GNNs are broadly used for vertex or graph classification and link prediction, where graph topology is the essential information for such task but the permutation prior is seldom concerned.
In wireless communications domain, GNNs have been designed for a variety of wireless tasks recently \cite{WCGNNsurvey2021}, aimed at learning wireless policies (i.e., the mapping from environment parameters to decision variables).
Enabling GNNs to exploit permutation prior is important for wireless communications due to the following three reasons, but the role of the topology prior is not obvious.
First, a wide range of wireless policies are with permutation properties \cite{Eisen2020,Multidimensional2023Liu}, because a large number of policies are defined on sets consisting of unordered elements \cite{SCJ} (e.g., a user set consisting of users whose ordering does not affect a precoding policy).
Second, the permutation prior is critical for supporting the scalability and size generalizability\footnote{Scalability means that a deep neural network (DNN) well trained in a small-scale system can still be well trained in large-scale systems with acceptable complexity \cite{lee2021graph}. Size generalizability  means that a DNN can perform well in unseen problem scales without re-training \cite{lee2021graph,shen2021graph}. } of the GNNs \cite{Eisen2020,shen2021graph,Guo2023Generalizable}, such that the GNNs are applicable to the wireless systems with large scales and with dynamic scales.
Third, it has been demonstrated that if the permutation property of the functions representable by a GNN does not match with the permutation property of the wireless policy to be learned (referred to as \emph{desired permutation property}), then the learning performance degrades \cite{guo2022learning} or the training complexity is high \cite{zhao2022understanding}.

In machine learning literature, it is often stated that GNNs are naturally permutation equivariant.
This is because the majority of GNNs considered therein learn over homogeneous graphs that are with one type of vertices and edges. These GNNs (called HomoGNNs) satisfy one-dimension (1D)-permutation equivariant (PE)/permutation invariant (PI) and joint PE/PI properties by using the same processing, pooling, and combining functions for all vertices (and edges).
However, many wireless policies have far more complicated permutation properties than 1D-PE/PI, say a combination of joint and nested PE/PI properties.
GNNs learning over heterogeneous graphs consisting of multiple types of vertices or edges (called  HetGNNs), can  satisfy these complicated properties only after a judicious design.

\subsection{Related Works}
\subsubsection{Designing GNNs for Wireless Communications}
GNNs have been used to optimize a variety of wireless problems, say power control/allocation, link scheduling and user association/scheduling, precoding and detection, in diverse system settings.

One motivation of adopting GNNs is to leverage the permutation prior for reducing training complexity \cite{shen2021graph, guo2022learning, Multidimensional2023Liu, Guo2023Generalizable, Wang2023Learning, zhao2022understanding}, enabling size generalizability \cite{shen2021graph, Jiang2021graph, Multidimensional2023Liu, Guo2023Generalizable, Chowdhury2021Unfolding, Wang2023Learning, Zhang2023GNN,Shelim2023Learning}, scalability \cite{shen2021graph} and learning performance \cite{Multidimensional2023Liu, MishraGNN2024}.

Another motivation is to leverage the topology prior  \cite{Chen2022graph, He2022graphtcom, Yang2024Graph,Multidimensional2023Liu,Chowdhury2021Unfolding, Zhang2023GNN, zhao2022understanding, Wang2023Learning}, for improving learning performance \cite{Chen2022graph, He2022graphtcom, Yang2024Graph}, enabling  size generalizability \cite{He2022graphtcom}, and reducing the training complexity \cite{Multidimensional2023Liu} of GNNs.

The aspects of designing a GNN for wireless tasks consist of modeling
a graph and designing the architecture.

{\em \textbf{{Designing Graphs:}}} Unlike many tasks in other domains such as graph learning where graph datasets are available, graphs for wireless problems need to be modeled by analyzing the problems. Constructing a graph includes defining vertices, edges, their features, actions and types. Most graphs are constructed heuristically.
For instance, a base station (BS) with multiple antennas \cite{Wang2023Learning} or an intelligent reflecting surface with multiple passive reflective elements \cite{Jiang2021graph} was defined as a vertex. Such a definition neglects the permutability of the antennas or reflective elements. As a consequence, the GNNs learning over the graphs are with high training complexity and are unable to be generalized to the numbers of these nodes.
In \cite{Multidimensional2023Liu}, a systematic method of constructing graphs for wireless policies to exploit permutation prior was provided. In particular, the sets in an optimization problem were first identified, from which the desired permutation property can be found, and then the vertices in a graph were defined according to the sets. However, this method is only applicable for the problems with independent sets, because the relation across multiple sets is neglected.

{\em \textbf{{Designing architecture of GNNs:}}}
GNNs can be divided into VertexGNNs and EdgeGNNs, where hidden representations of every vertex and every edge  are respectively updated. They are usually applied for the problems when the decision variables (called actions) are defined on vertices and on edges, respectively. Then, the hidden representations in the last layer can be taken as the actions without the need of a readout function in the output layer.
In each layer of a GNN, the hidden representation
of every vertex or edge in a graph is updated by first extracting
information from neighboring vertices or edges with a processor,
then aggregating the information with a pooling function,
and finally combining the information with the representation
of the vertex or edge itself by a combiner.

In the literature of wireless communications, the architecture design of a GNN focuses on selecting its
core components (i.e., processors, pooling functions, combiners, and read-out functions). For example, in the considered VertexGNNs, linear processors were used in \cite{lee2021graph, guo2022learning, Chowdhury2021Unfolding, Zhang2023GNN, MishraGNN2024, He2022graphtcom, Yang2024Graph, Zhangmodelgnn2025}, fully-connected neural networks (FNNs) and convolutional neural networks were respectively used as processors in \cite{shen2021graph, Wang2023Learning, Jiang2021graph} and in \cite{Chen2022graph}, logarithmic functions with trainable parameters were designed as processors in \cite{Shelim2023Learning}. In the considered EdgeGNNs, linear processors were used in \cite{Eisen2020, zhao2022understanding}, and FNNs were used as processors in \cite{Multidimensional2023Liu, Guo2023Generalizable, Wang2023Learning}. The pooling
functions are usually selected from summation, mean, and maximization functions, and combiners are often selected as FNNs.

\subsubsection{Design DNNs with Desired Permutation Properties}
Most earlier works consider HomoGNNs, while recent research efforts started to design HetGNNs. In \cite{Chowdhury2021Unfolding, shen2021graph, Zhang2023GNN,Shelim2023Learning, Zhangmodelgnn2025}, HomoGNNs were designed to satisfy desired permutation properties. In \cite{guo2022learning, zhao2022understanding, Guo2023Generalizable, Wang2023Learning, Jiang2021graph, Multidimensional2023Liu,MishraGNN2024}, HetGNNs were designed to satisfy partial \cite{Wang2023Learning, Jiang2021graph} or all \cite{guo2022learning, zhao2022understanding, Guo2023Generalizable, Multidimensional2023Liu,MishraGNN2024} desired permutation properties. All these works employ a ``first-design-then-prove'' approach, which first design the architectures of GNNs and then prove that the designed GNNs are with permutation properties.

In addition to GNNs, permutation equivalent neural networks (PENNs) exhibit permutation properties, which were designed by introducing parameter sharing into FNNs \cite{zaheer2017deep, SEquivariance,maron2019universality,hartford2018deep, finzi2021practical}.
In \cite{zaheer2017deep, SEquivariance,maron2019universality,hartford2018deep}, the parameter sharing was designed by enumerating all possible permutations of the elements in the input and output vectors of a FNN, which is rather involved and hence is hard to be used for many wireless policies with complex permutation properties. In \cite{finzi2021practical}, a low complexity method was proposed, where an optimization problem was formulated to find the weight matrix with parameter sharing and an iterative algorithm was provided to solve the problem, but the optimized weight matrix is inaccurate.

It has been noticed that there are two aspects that enable a HetGNN to satisfy desired permutation property: designing an appropriate graph \cite{Multidimensional2023Liu} and designing the architecture of HetGNN \cite{guo2022learning, Wang2023Learning}.
However, these aspects have not been well-addressed so far.

\emph{\textbf{Challenge 1: How to systematically construct a graph for a HetGNN to exploit permutation prior?}} The constructed graph for a wireless problem may not be unique. Different graphs for a problem may not entail the difference in learning performance and efficiency, depending on the desired permutation property. For instance, for learning link scheduling policy or power control policy in device-to-device (D2D) communications, each transceiver was defined as a vertex in \cite{lee2021graph,shen2021graph,He2022graphtcom} and the transmitters and receivers were defined as two types of vertices in \cite{PY}. The GNNs learning over the graphs constructed in both ways perform the same, given that the policies are with joint PE property.

However, when the desired permutation property is induced by multiple sets that are related in terms of permutability, the property cannot be satisfied by a GNN no matter how the GNN is designed if the graph is not well modeled.

The difficulties of modeling graph to exploit permutation prior lie in the following two aspects.
First, the mapping from entities (say users) to the vertices of graph cannot be expressed as a closed-form expression. Second, the relations among vertices is complicated (say serving and competition) and the mapping from the vertices to the their relations (i.e., edges) also cannot be expressed as a closed-form expression. Consequently, it is hard to employ mathematical tools to design the mappings.

\emph{\textbf{Challenge 2: How to systematically design the architecture of a HetGNN to exploit permutation prior?}}
Even if the graph has been well-constructed for a wireless problem, the design of a HetGNN to satisfy desired permutation property is still non-trivial.

In machine learning field, several HetGNNs have been proposed \cite{bing2023heterogeneous}, but none of these HetGNNs were designed for satisfying permutation properties.
For example, a VertexGNN was designed in \cite{schlichtkrull2018modeling}, which updates the hidden representation of every vertex according to the types of edges during aggregation.
In \cite{hu2020heterogeneous}, a VertexGNN was designed that updates the representation of a vertex according to the type of the vertex and the types of its neighbouring vertices and edges, which however still cannot satisfy desired permutation property if the graph is not well-constructed for a problem.

While HetGNNs have been designed to satisfy all or partial permutation properties of wireless  policies \cite{guo2022learning, zhao2022understanding, Guo2023Generalizable, Wang2023Learning, Jiang2021graph, Multidimensional2023Liu,MishraGNN2024}, the ``first-design-then-prove'' approach is lack of a design principle, which is hard to exploit permutation prior for the problems with more than two sets.

\subsection{Motivation and Contributions}
In this paper, we strive to address the two challenges by proposing methods for designing heterogeneous graphs and HetGNNs to satisfy desired permutation properties.
To demonstrate how to use the methods, we take two representative wireless problems as examples, which are power allocation in multi-cell-multi-user systems and hybrid precoding in multi-user multi-input-multi-output (MU-MIMO) mmWave communications, respectively. The first policy satisfies joint and nested permutation property, and the second policy satisfies joint, independent and nested permutation property.

The major contributions are summarized as follows.

\begin{itemize}
  \item We find the connection among the \emph{sets} that have been noticed in wireless problems so far, the \emph{types of edges} in a graph, and the \emph{joint and nested permutation properties}. Based on the finding,
      we propose a method for systematically constructing graphs for learning wireless policies, such that the complicated permutation properties can be satisfied by HetGNNs with judiciously designed architectures.
  \item  We find and prove sufficient conditions for processors, pooling functions, and combiners to satisfy, which suggests a systematical method of designing the architectures of HetGNNs. The proposed method is easy-to-use, with which the designed HetGNNs learning over well-modeled graphs satisfy desired permutation properties.
  \item We use simulations to show the impact of exploiting permutation prior and compare with the impact of harnessing topology prior, on both training complexity and size generalization ability. 
\end{itemize}

The rest of the paper is organized as follows.
In Section \ref{per-prior}, we introduce notions regarding permutation prior and the approaches to exploit the prior. In Section \ref{section:graph-gnn} and  Section \ref{section:permutation}, we propose the methods to construct graphs and design HetGNNs to satisfy desired permutation properties, respectively.
In Section \ref{section:V}, we summarize the procedure of systematically designing graphs and HetGNNs to exploit permutation prior and compare the proposed methods with existing methods.
Simulation results are provided in Section \ref{section: simulation}, and conclusions are provided in Section \ref{conclu}.

{\em Notations:} $(\cdot)^\mathsf{T}$, $(\cdot)^\mathsf{H}$,  ${\rm Tr}(\cdot)$, and $||\cdot||_{\sf F}$ denote the transpose, Hermitian transpose, trace, and Frobenius norm of a matrix, respectively. ${x}_{i}$ and ${X}_{ij}$  are the $i$-th element in a vector ${\bf x}$, and the element in the $i$-th row and $j$-th column of a matrix ${\bf X}$, respectively. $\otimes$ denotes Kronecker product. $|\cdot|$ denotes magnitude. ${\bf X} = {\rm diag}({\bf x})$ denotes a diagonal matrix whose diagonal elements are the elements in ${\bf x}$ while ${\bf X} = {\rm diag}({\bf X}_{1}, \cdots, {\bf X}_{K})$ denotes a block diagonal matrix.
${\bf I}_{K} \in \mathbb{R}^{K \times K}$ denotes identity matrix, and $\{0, 1\}^{M \times M}$ denotes a $M \times M$ matrix whose elements are either zero or one. $\pi(\cdot)$ stands for the permutation operation on the indices of the elements of a set, where $\pi(i)=j$ means that the element initially with  index ``$i$'' is with index ``$j$'' after re-ordering.
${\bf \Pi}$ is a matrix that permutes the rows of a vector or a matrix.

\section{Permutation Prior} \label{per-prior}
In this section, we formally introduce several notions and explain them with concrete wireless problems. Then, we provide the approaches of designing DNNs to exploit permutation prior.


\subsection{Wireless Policy, Permutation Property, and Sets}
\label{relation-ppsp}
A {\bf wireless policy} is a mapping from the concerned environment parameters to the decision variables, which is a multivariate function and can be obtained from an optimization problem. For example, the baseband precoding policy in multi-user multi-input-single-output (MU-MISO) systems can be defined as the mapping from channel matrix ${\bf H}  \in \mathbb{C}^{K \times N_t}$ to precoding matrix ${\bf V}  \in \mathbb{C}^{K \times N_t}$, denoted as ${\bf V}={f_{\tt B}}({\bf H})$, where $N_t$ and $K$ are respectively the numbers of antennas at the BS and the user equipments (UEs).

 A multivariate function has {\bf permutation
property} if its output is equivalent or invariant
to the permutation of its input, which is called a \emph{permutable function}. For example, a function ${\bf y}=f({\bf x})$ is 1D-PE to ${\bf x}$ if ${\bm \Pi}{\bf y} = f({\bm \Pi}{\bf x})$, and is 1D-PI to ${\bf x}$ if ${\bf y} = f({\bm \Pi}{\bf x})$, where ${\bf y}$ and ${\bf x}$ are respectively the vectors of output and input variables. The 1D permutation property (i.e., 1D-PE or 1D-PI  property) is the permutation prior for designing the DNN to learn $f(\cdot)$.

Permutation properties widely exist in wireless policies. This is because a function defined on a set is permutable \cite{zaheer2017deep}, while these policies are obtained from the wireless problems defined on sets \cite{SCJ}. For example, the baseband precoding policy ${\bf V}={f_{\tt B}}({\bf H})$ are defined on UE set and antenna set.
Since the functions on different kinds of sets have different permutation properties, next we introduce the sets that have been found in wireless problems and the resulting permutation properties.

\begin{figure}[!htb]
\centering
\subfigure[Normal set]{
\includegraphics[width=0.15\textwidth]{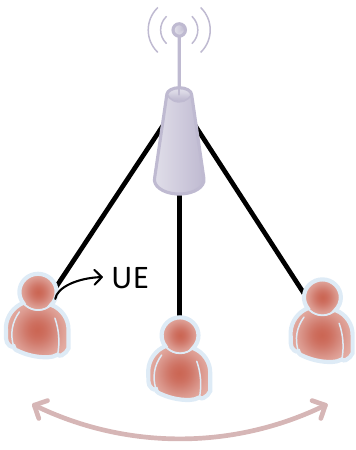}
\label{fig:normal}
} \hspace{2mm}
\subfigure[Nested set]{
\includegraphics[width=0.2\textwidth]{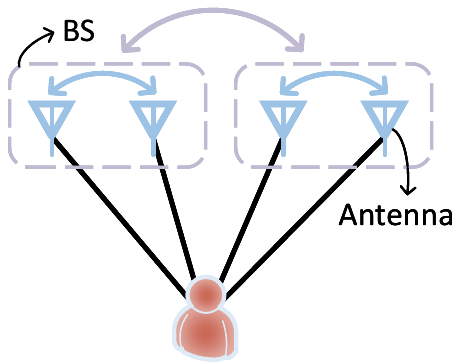}
\label{fig:nested}
} \\
\subfigure[Related sets]{
\includegraphics[width=0.15\textwidth]{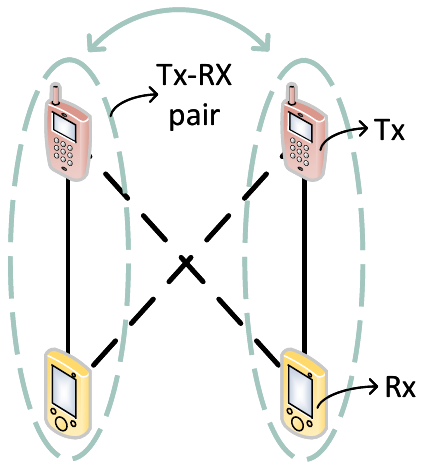}
\label{fig:related}
} \hspace{3mm}
\subfigure[Non-related sets]{
\includegraphics[width=0.16\textwidth]{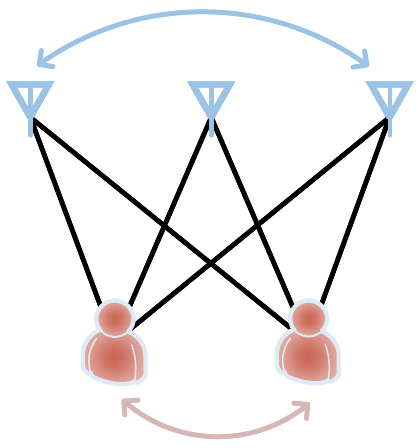}
\label{fig:non-related}
}
\caption{Examples of sets and their relations}
\label{fig: sets}
\end{figure}

A {\bf set} is composed of multiple elements (say UEs), which can either be a normal set or a nested set, depending on whether the set consists of elements or subsets.\footnote{\scriptsize In \cite{Pinter2014}, the normal set only with elements but not with subsets is called a set, and the nested set is referred to as a family of sets.}
\vspace{-0.1mm}
\begin{itemize}
\item A {\bf normal set} consists of elements that can be re-ordered arbitrarily. A function defined on a normal set exhibits 1D permutation property \cite{zaheer2017deep, maron2019universality}, joint or high-order permutation property \cite{maron2019universality}, depending on the dimension of input variable of the function. If the input of the function is a vector, matrix, or high-order tensor, meanwhile the variable corresponding to an element in the set is a scalar, then the function has the 1D, joint, or high-order permutation properties.

    For example, the \emph{power allocation problem in an interference-free multi-UE system} consists of a normal set of UEs, as illustrated in Fig. \ref{fig:normal}.
    The power allocation policy defined on the set is ${\bf p}=g_{\tt MU}({\bf h})$, where ${\bf p}=[p_1, p_2, p_3]^{\sf T}$ and ${\bf h}=[h_1, h_2, h_3]^{\sf T}$ when there are three UEs, $p_k$ and $h_k$ are respectively the allocated power and the channel coefficient of UE$_k$.

     When the UEs are re-ordered (i.e., their indices are permuted with $\pi(\cdot)$), the elements in ${\bf p}$ and ${\bf h}$ are permuted. For instance, when $[\pi(1), \pi(2), \pi(3)]^{\sf T}=[2, 1, 3]^{\sf T}$, ${\bf p}$ is permuted as ${\bf p'}=[p_{\pi(1)}, p_{\pi(2)}, p_{\pi(3)}]^{\sf T}=[p_2, p_1, p_3]^{\sf T}$ and ${\bf h}$ is permuted as ${\bf h'}=[h_{\pi(1)}, h_{\pi(2)}, h_{\pi(3)}]^{\sf T}=[h_2, h_1, h_3]^{\sf T}$. Considering the relation between permutation matrix and permutation operation,
${\bf p'}$ and ${\bf h'}$ can also be respectively expressed as  ${\bf p'}={\bm \Pi}_{1} {\bf p}$ and ${\bf h'}={\bm \Pi}_{1} {\bf h}$.\footnote{Here, \scriptsize ${\bm \Pi}_{1}=$ $\def\arraystretch{0.7}\begin{bmatrix}
0 & 1 & 0 \\
1 & 0 & 0 \\
0 & 0 & 1 \\
\end{bmatrix}$, which can be obtained from ${\bm \Pi}_{1}[1, 2, 3]^{\sf T}=[\pi(1), \pi(2), \pi(3)]^{\sf T}=[2, 1, 3]^{\sf T}$.} Because the power allocation policy remains unchanged after the permutation, it satisfies ${\bm \Pi}_{1}{\bf p}=g_{\tt MU}({\bm \Pi}_{1}{\bf h})$, i.e., the 1D-PE property.


\item A {\bf nested set} consists of unordered subsets and each subset consists of unordered elements, where the subsets and the elements in each subset can be re-ordered arbitrarily.
    A function defined on a nested set has nested permutation property \cite{SEquivariance}.

    For example, a precoding problem in a \emph{coordinated multi-point with joint transmission (CoMP-JT) system where multi-antenna BSs serves a single-antenna UE} consists of a nested set: \{\{antennas at BS$_1$\}, \{antennas at BS$_2$\}, $\cdots$\}, as illustrated in Fig. \ref{fig:nested}. The antennas at all BSs constituting the nested set is because the maximal power constraint at each BS.
    The precoding policy is ${\bf v} =g_{\tt CoMP}({\bf h})$, where ${\bf v}=[v_{1_1}, v_{2_1}, v_{1_2}, v_{2_2}]$ and ${\bf h}=[h_{1_1}, h_{2_1}, h_{1_2}, h_{2_2}]$ when there are two BSs and each BS is equipped with two antennas, $v_{i_j}$ and $h_{i_j}$ are respectively the precoding weight of the $i$-th antenna at BS$_j$ (called AN$_{i_j}$ for short) and the channel coefficient from the antenna to the UE.

    When the BSs are reordered, the precoding vector ${\bf v}$ is permuted as ${\bf v'}=[v_{1_{\pi(1)}}, v_{2_{\pi(1)}}, v_{1_{\pi(2)}}, v_{2_{\pi(2)}}]=[v_{1_2}, v_{2_2}, v_{1_1}, v_{2_1}]={\bm \Pi}_{2}{\bf v}$, and the channel vector ${\bf h}$ is permuted as ${\bf h'}=[h_{1_{\pi(1)}}, h_{2_{\pi(1)}}, h_{1_{\pi(2)}}, h_{2_{\pi(2)}}]=[h_{1_2}, h_{2_2}, h_{1_1}, h_{2_1}]={\bm \Pi}_{2}{\bf h}$.\footnote{\scriptsize ${\bm \Pi}_{2}={{\bm \Pi}}_{{\tt BS}} \otimes {\bf I}_2 = $ $\def\arraystretch{0.7}\begin{bmatrix}
0 & 0 & 1 & 0 \\
0 & 0 & 0 & 1 \\
1 & 0 & 0 & 0 \\
0 & 1 & 0 & 0
\end{bmatrix}$, where $\def\arraystretch{0.7}{{\bm \Pi}}_{{\tt BS}}=\begin{bmatrix}
0 & 1 \\
1 & 0
\end{bmatrix}$ represents the permutation of two BSs and is obtained from ${{\bm \Pi}}_{{\tt BS}}[1, 2]^{\sf T}=[\pi(1), \pi(2)]^{\sf T}=[2, 1]^{\sf T}$. Since each BS is with two antennas, ``$\otimes {\bf I}_2$'' is used to obtain the permutations of all antennas when the BSs are re-ordered.}

The antennas at each BS can also be re-ordered.
We take re-ordering the antennas at BS$_1$ as an example. When the antennas are re-ordered, ${\bf v}$ is permuted as ${\bf v''}=[v_{\pi(1)_{1}}, v_{\pi(2)_{1}}, v_{1_{2}}, v_{2_{2}}]=[v_{2_{1}}, v_{1_{1}}, v_{1_{2}}, v_{2_{2}}]={\bm \Pi}_{3}{\bf v}$ and ${\bf h}$ is permuted as ${\bf h''}=[h_{\pi(1)_{1}}, h_{\pi(2)_{1}}, h_{1_{2}}, h_{2_{2}}]=[h_{2_{1}}, h_{1_{1}}, h_{1_{2}}, h_{2_{2}}]={\bm \Pi}_{3}{\bf h}$.\footnote{\scriptsize $\def\arraystretch{0.7}{\bm \Pi}_{3}={\rm diag}({{\bm \Pi}}_{{\tt AN, {\tt BS}_1}}, {\bf I}_2)=\begin{bmatrix}
0 & 1 & 0 & 0 \\
1 & 0 & 0 & 0 \\
0 & 0 & 1 & 0 \\
0 & 0 & 0 & 1
\end{bmatrix}$, where $\def\arraystretch{0.7}{{\bm \Pi}}_{{\tt AN, {\tt BS}_1}}=\begin{bmatrix}
0 & 1 \\
1 & 0
\end{bmatrix}$ represents the permutation of two antennas at BS$_1$ and is obtained from ${{\bm \Pi}}_{{\tt AN, {\tt BS}_1}}[1, 2]^{\sf T}=[\pi(1), \pi(2)]^{\sf T}=[2, 1]^{\sf T}$, and ${\bf I}_2$ indicates that the antennas at BS$_2$ are not re-ordered.
}

When the BSs and the antennas at each BS (say BS$_1$) are re-ordered simultaneously, the reordering can be  equivalently decomposed into first re-ordering the antennas at BS$_1$ and then re-ordering the BSs, or first re-ordering the BSs and then re-ordering the antennas at BS$_1$. As a result, ${\bf v}$ and ${\bf h}$ are respectively permuted as ${\bm \Pi}_{2}{\bm \Pi}_{3}{\bf v}$ and ${\bm \Pi}_{2}{\bm \Pi}_{3}{\bf h}$, or ${\bm \Pi}_{3}{\bm \Pi}_{2}{\bf v}$ and ${\bm \Pi}_{3}{\bm \Pi}_{2}{\bf h}$.
The equivalency of the two reorderings comes from the fact that ${\bm \Pi}_{2}{\bm \Pi}_{3}={\bm \Pi}_{3}{\bm \Pi}_{2}$. By substituting the expressions of ${\bm \Pi}_{2}$ and ${\bm \Pi}_{3}$ into ${\bm \Pi}_{2}{\bm \Pi}_{3}{\bf v}$ and ${\bm \Pi}_{2}{\bm \Pi}_{3}{\bf h}$, we can obtain that the precoding policy satisfies the nested permutation property: $({{\bm \Pi}}_{{\tt BS}} \otimes {\bf I}_2){\rm diag}({{\bm \Pi}}_{{\tt AN, {\tt BS}_1}}, {\bf I}_2) {\bf v}=g_{\tt CoMP}(({{\bm \Pi}}_{{\tt BS}} \otimes {\bf I}_2){\rm diag}({{\bm \Pi}}_{{\tt AN, {\tt BS}_1}}, {\bf I}_2){\bf h})$.
\end{itemize}

{\bf Multiple sets} may either be {related or not related} in terms of permutability.
\begin{itemize}
\item Several sets are {\bf related} if the elements in these sets should be re-ordered together. The function defined on related sets exhibits joint permutation property  \cite{maron2019universality}.

    For example, a \emph{link scheduling or power control problem in D2D communications} consists of two related normal sets: transmitter (Tx) set and receiver (Rx) set, where the Tx and Rx in each transceiver pair should be re-ordered together as illustrated in Fig. \ref{fig:related}. The policy defined on the sets is ${\bm \rho} = g_{\tt D2D}({\bf H})$, where ${\bm \rho}=[{\rho}_1, {\rho}_2]$ and ${\bf H}=[[h_{11}, h_{12}]^{\sf T}, [h_{21}, h_{22}]^{\sf T}]$ when there are two transceiver pairs, ${\rho}_i$ is the decision variable of the $i$-th transceiver pair, and $h_{ij}$ is the channel gain from the Tx in the $i$-th pair to the Rx in the $j$-th pair.

    When the transceiver pairs are re-ordered, the decision vector ${\bm \rho}$ is permuted as ${\bm \rho}'=[{\rho}_{\pi(1)}, {\rho}_{\pi(2)}]=[{\rho}_2, {\rho}_1] ={\bm \Pi}_{4}{\bm \rho}$, and the channel matrix ${\bf H}$ is permuted as ${\bf H}'=[[h_{\pi(1)\pi(1)}, h_{\pi(1)\pi(2)}]^{\sf T}, [h_{\pi(2)\pi(1)}, h_{\pi(2)\pi(2)}]^{\sf T}]=[[h_{22}, h_{21}]^{\sf T}, [h_{12}, h_{11}]^{\sf T}]={\bm \Pi}_{4}{\bf H}{\bm \Pi}_{4}^{\sf T}$. Then, the policy satisfies the joint PE property: ${\bm \Pi}_{4}{\bm \rho} = g_{\tt D2D}({\bm \Pi}_{4}{\bf H}{\bm \Pi}_{4}^{\sf T})$.\footnote{\scriptsize $\def\arraystretch{0.7}{\bm \Pi}_{4}=\begin{bmatrix}
0 & 1 \\
1 & 0
\end{bmatrix}$, $\def\arraystretch{0.7}{\bm \Pi}_{5}=\begin{bmatrix}
0 & 0 & 1\\
1 & 0 & 0 \\
0 & 1 & 0
\end{bmatrix}$, and $\def\arraystretch{0.7}{\bm \Pi}_{6}=\begin{bmatrix}
0 & 1 \\
1 & 0
\end{bmatrix}$.
}

\vspace{1mm}{\bf \emph{Remark 1}:}
    This example problem can also be regarded as consisting of a normal set, where each element in the set is a transceiver pair \cite{lee2021graph, shen2021graph, PY}.
    In this way, the policy defined on the set also satisfies the joint PE property.
    However, regarding multiple related sets as a normal set will result in incorrect permutation property if each set is a nested set.
    For generality, we do not regard multiple related sets as a normal set.

\item Several sets are {\bf not related} if the elements in different sets can be permuted independently. The function defined on multiple unrelated sets has multi-dimension independent permutation property \cite{hartford2018deep}.

    For example, the \emph{baseband precoding problem for MU-MISO system} consists  of two normal sets: UE set and antenna set, as illustrated in Fig. \ref{fig:non-related}. The policy defined on the sets is ${\bf V}=f_{\tt B}({\bf H})$, where ${\bf V}=[[v_{11}, v_{12}, v_{13}]^{\sf T}, [v_{21}, v_{22}, v_{23}]^{\sf T}]$ and ${\bf H}=[[h_{11}, h_{12}, h_{13}]^{\sf T}$, $[h_{21}, h_{22}, h_{23}]^{\sf T}]$ when $K=2$ and $N_t=3$, ${v}_{ij}$ and $h_{ij}$ are the precoding and channel coefficient from the $j$-th antenna to UE$_i$, respectively.

    The elements in the UE set can be re-ordered arbitrarily no matter how the elements in the antenna set are re-ordered. When the UEs and the antennas are re-ordered, the precoding matrix ${\bf V}$ and channel matrix ${\bf H}$ are permuted accordingly, say ${\bf V'}=[[v_{23}, v_{21}, v_{22}]^{\sf T}, [v_{13}, v_{11}, v_{12}]^{\sf T}]={\bm \Pi}_{5}{\bf V}{\bm \Pi}_{6}$ and ${\bf H'}=[[h_{23}, h_{21}, h_{22}]^{\sf T}, [h_{13}, h_{11}, h_{12}]^{\sf T}]={\bm \Pi}_{5}{\bf H}{\bm \Pi}_{6}$. Then, the policy satisfies ${\bm \Pi}_{5}{\bf V}{\bm \Pi}_{6}=f_{\tt B}({\bm \Pi}_{5}{\bf H}{\bm \Pi}_{6})$, i.e., the two-dimension (2D)-PE property.
\end{itemize}

\subsection{Design DNNs for Exploiting Permutation Prior}
Denote a wireless policy as $p(\cdot)$, whose permutation property is induced by the sets of the corresponding problem, i.e., the \emph{desired permutation property}.

Denote the learned policy by a DNN as ${\hat p}(\cdot)$, which is a function selected from the function family represented by the DNN that can minimize the loss function.

The permutation prior is a property of $p(\cdot)$ in terms of permutability, which can be used for designing a DNN to learn the policy efficiently. \emph{Exploiting the permutation prior} means that the desired permutation property induced by all the sets of $p(\cdot)$ is harnessed for designing the DNN. If a DNN is designed only by considering the permutation induced by some but not all sets of $p(\cdot)$, the permutation property of ${\hat p}(\cdot)$ does not match with the desired permutation property.

Both PENNs and GNNs can be designed to exploit permutation prior, but with different approaches.

\subsubsection{Approach for PENNs}
The approach to design a PENN for exploiting permutation prior is with the following procedure: \vspace{-1mm}
\begin{itemize}
\item {\em S$_{\tt PENN}$-1:} Identify the sets of a wireless problem.
\item {\em S$_{\tt PENN}$-2:} Analyze the permutation property of $p(\cdot)$ according to the sets.
\item {\em S$_{\tt PENN}$-3:} Introduce parameter sharing into FNNs according to the permutation property.
\end{itemize}
\vspace{-1mm}

Systematical methods for addressing S$_{\tt PENN}$-3 have been proposed in \cite{zaheer2017deep, SEquivariance,maron2019universality,hartford2018deep, finzi2021practical} for functions with given permutation properties. The issues in S$_{\tt PENN}$-1 and S$_{\tt PENN}$-2 depend on the concerned problems, which have been discussed in \cite{Multidimensional2023Liu}.

\subsubsection{Approach for GNNs}
While similar approach can be used, GNNs can be designed with a much simpler approach. This can be achieved by connecting the sets of a policy with a graph and finding the components in a GNN that affect the permutability.

A graph consists of vertices and edges, where each vertex or each edge may be with a feature and an action. In wireless problems, a vertex may be a node (say a UE) or a virtual entity (say a data stream), an edge connects interactive vertices, a feature can be the environment parameter (say channel), and an action is the decision variable.

Denote the function defined on a graph $\mathcal{G}$ as ${\tilde p}(\mathcal{G})$, which is a mapping from $\mathcal{G}$ to the decision variables. The permutation property of ${\tilde p}(\mathcal{G})$ is induced by the graph.
If the graph can reflect the permutability of the sets of $p(\cdot)$, then ${\tilde p}(\mathcal{G})$ will be with the same permutation property as $p(\cdot)$.

A GNN can represent a family of functions defined on a graph $\mathcal{G}'$. Denote the learned policy by a GNN as ${\hat p}(\mathcal{G}')$.
If the GNN learns over $\mathcal{G}$ meanwhile its architecture can be appropriately designed, the permutation property of ${\hat p}(\mathcal{G})$ is the same as the permutation property of ${\tilde p}(\mathcal{G})$.

\begin{figure*}[!htb]
\centering
\begin{minipage}[c]{1\textwidth}
\centering
\includegraphics[scale=0.35]{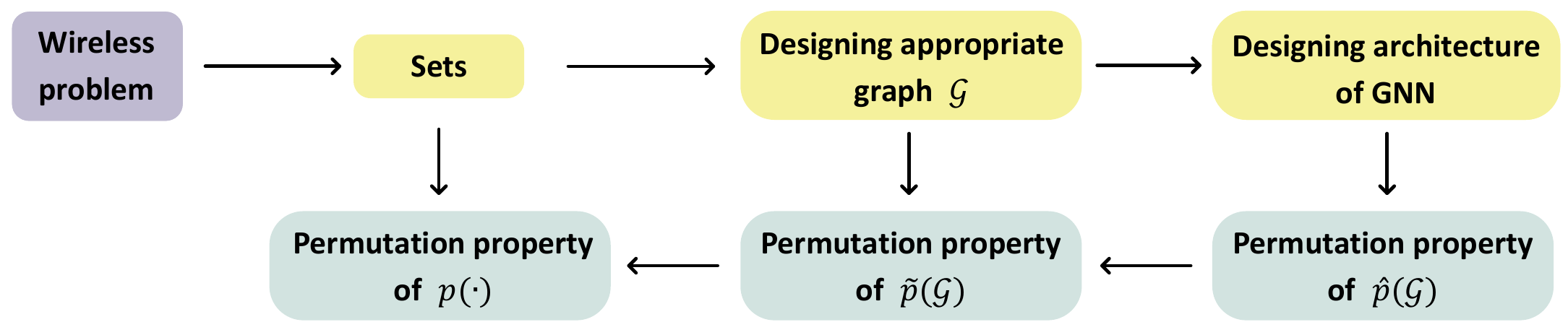}
\end{minipage}
\caption{Schematic diagram of designing GNN to exploit permutation prior}
\label{fig:methods} \vspace{-4mm}
\end{figure*}

The approach to design a GNN for exploiting permutation prior is with the following procedure:
\vspace{-1mm}
\begin{itemize}
\item {\em S$_{\tt GNN}$-1:} Identify the sets of a wireless problem (same as S$_{\tt PENN}$-1).
\item {\em S$_{\tt GNN}$-2:} Construct a graph $\mathcal{G}$ such that ${\tilde p}(\mathcal{G})$ exhibits the same permutation property as $p(\cdot)$.
\item {\em S$_{\tt GNN}$-3:} Design the architecture of the GNN such that ${\hat p}(\mathcal{G})$ exhibits the same permutation property as ${\tilde p}(\mathcal{G})$.
\end{itemize}
\vspace{-1mm}

The procedure is shown in Fig. \ref{fig:methods}.

So far, there is still lack of systematical methods of modeling the heterogeneous graphs and designing the architecture of HetGNNs to satisfy complicated permutation properties. In the following sections, we propose methods to address the issues in S$_{\tt GNN}$-2 and S$_{\tt GNN}$-3.



\section{Constructing Heterogeneous Graphs} \label{section:graph-gnn}
In this section, we show the limitation of an existing method to construct graphs from the sets of wireless problems. Then, we propose a method for constructing a graph, such that the function on the graph is  with desired permutation property. We proceed to use two wireless problems to demonstrate how to apply the new method.

\subsection{Methods to Construct Graph}
A heterogeneous graph consists of multiple types of vertices or edges.

To enable the function defined on graph ${\tilde p}(\cdot)$ to exhibit the same permutation property as a wireless policy $p(\cdot)$, we should construct \emph{the graph that ${\tilde p}(\cdot)$ is defined on} from \emph{the sets that $p(\cdot)$ is defined on}.

The desired permutation properties having been found in wireless communications consist of independent, joint, nested permutation properties or their combinations. Constructing a heterogeneous graph $\mathcal{G}$ such that ${\tilde p}(\mathcal{G})$ satisfies a combination of these properties is non-trivial.

\subsubsection{Limitation of an Existing Method}
Constructing a graph for learning a wireless policy is composed of defining vertices, edges, features, actions, and the types of vertices and edges.

In \cite{Multidimensional2023Liu}, a method to  construct graphs from unrelated sets was provided, which consists of the following two steps.

{({\tt S1})} After identifying the sets of a wireless problem, the elements in each set are defined as the vertices of each type.
The features and actions of the vertices are defined according to the input and output tensors of the resulting policy.

{({\tt S2})} Define edges as well as their features and actions according to the input and output tensors.

The method has two limitations as follows.

{\em \textbf{{Limitation 1:}}}
The types of edges are not defined, which do not affect the permutation property when multiple sets are not related.
However, the types of vertices cannot reflect the permutability across related sets or the permutability between a nested set and its subsets. Hence, only defining the types of vertices does not ensure that the permutation property of ${\tilde p}(\cdot)$ is the same as the permutation property of $p(\cdot)$.

\begin{figure}[!htb]
\centering
\subfigure[One type of edges]{
\includegraphics[width=0.14\textwidth]{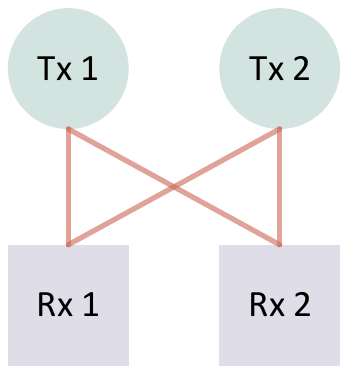}
\label{fig:e1-lsj}
} \hspace{3mm}
\subfigure[Two types of edges]{
\includegraphics[width=0.14\textwidth]{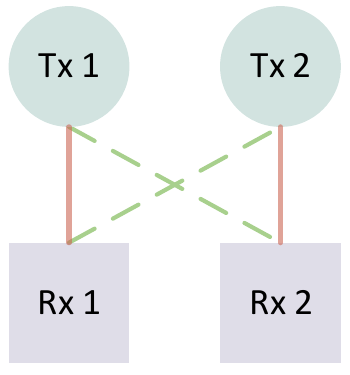}
\label{fig:e1}
} \\
\subfigure[One type of edges]{
\includegraphics[width=0.18\textwidth]{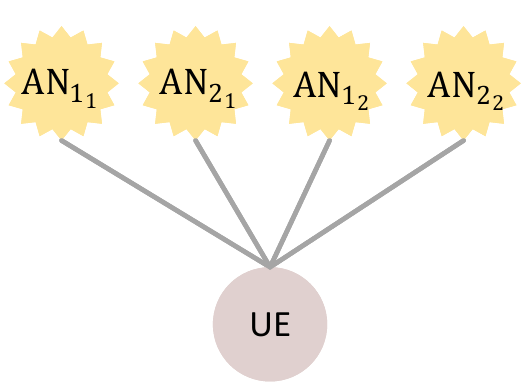}
\label{fig:e2-lsj}
}
\subfigure[Two types of edges]{
\includegraphics[width=0.18\textwidth]{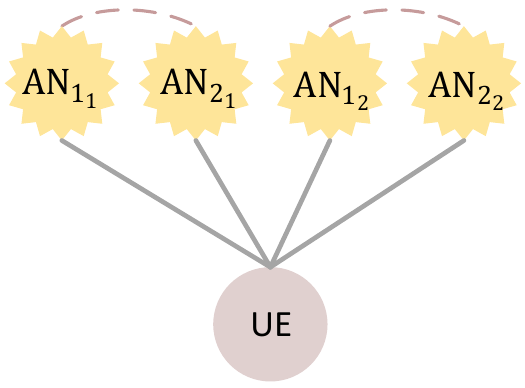}
\label{fig:e2}
}
\caption{Graphs constructed for two simple policies, (a) and (b) are graphs for a D2D communication system with two Tx-Rx pairs, (c) and (d) are graphs for a CoMP-JT system with two BSs, the vertices or the edges with the same color belong to the same type}
\label{fig: ModelingGraph}
\end{figure}

{\em \textbf{{Example of Limitation 1:}}}
We take the link scheduling or power control policy in D2D communications ${\bm \rho} =g_{\tt D2D}({\bf H})$ in subsection \ref{relation-ppsp} as an example, which is defined on two related sets, i.e., Tx set and Rx set, where the Tx and Rx in each transceiver pair should be permuted jointly.
According to the existing method, the graph is with two types of vertices, and the actions are defined on Tx vertices, while the graph can either be with one or two types of edges.
If all the edges are with the same type, as illustrated in Fig. \ref{fig:e1-lsj}, then the Tx vertices or the Rx vertices can be permuted arbitrarily, i.e., the Tx and Rx vertices in a transceiver pair are not permuted jointly.
As a result, the function defined on the graph satisfies ${\bm \Pi}_{\tt Tx}{\bm \rho}={\tilde g}_{\tt D2D}({\bm \Pi}_{\tt Tx}{\bf H}{\bm \Pi}_{\tt Rx}^{\sf T})$, which is not the joint PE property. This issue can be avoided by constructing a homogeneous graph for this policy as in \cite{lee2021graph,shen2021graph}, but is unable to be circumvented for other wireless policies (say the hybrid precoding policy to be discussed later).

{\em \textbf{{Limitation 2:}}}
The edges are only defined based on the dimensions of the input and output tensors of a policy, disregarding the interaction among the actions due to practical constraints (say power or hardware constraints). Then, this also does not ensure that the permutation property of ${\tilde p}(\cdot)$ is the same as the permutation property of $p(\cdot)$.

{\em \textbf{{Example of Limitation 2:}}}
We take the precoding policy in the CoMP-JT system ${\bf v} = g_{\tt CoMP}({\bf h})$ in subsection \ref{relation-ppsp} as an example, which is defined on a nested set.
By using the existing method, the graph constructed for this policy is illustrated in Fig \ref{fig:e2-lsj}, where the actions (i.e., ${\bf v}$) are defined on the AN vertices and the features (i.e., ${\bf h}$) are defined on edges. The interaction among the actions due to the maximal power constraint at each BS is not reflected on the graph.
By using ${{\bm \Pi}}_{{\tt AN}, {\tt All}}$ to represent the permutation of all the four antennas at the two BSs, the function defined on this graph satisfies ${{\bm \Pi}}_{{\tt AN}, {\tt All}} {\bf v} = {\tilde g}_{\tt CoMP}({{\bm \Pi}}_{{\tt AN}, {\tt All}} {\bf h})$, which is not the same as the permutation property of this precoding policy defined on the nested set.

\subsubsection{Proposed Method}
Considering that related sets and nested sets widely exist in wireless communications, we propose a method for constructing a graph to avoid the limitations. The rationale behind the method is: \emph{the edges and the types of edges can capture the dependency among sets in terms of permutability} (i.e., the relation between the elements of two related  sets, and the relation between a nested set and its subsets), \emph{after the elements in different sets are defined as the vertices with different types}.

According to the new method, a graph is constructed  with the four steps as follows.
\vspace{-1mm}
\begin{itemize}
\item {({\tt S1$_{\tt new}$})} Define the vertices, their types and features or actions as {({\tt S1})}.

\item {({\tt S2$_{\tt new}$})} Define the edges, their features and actions as {({\tt S2})}.
Then, we define the types of edges, by \emph{examining which vertices are permutable} and \emph{which edges can be permuted according to the permutability of vertices}. Only the edges able to be permuted with the permutable vertices are defined as the same type.

\item {({\tt S3$_{\tt new}$})} When a policy is defined on sets containing nested set, examine if the function defined on the constructed graph satisfies the desired permutation property. If yes, then the graph construction is completed. Otherwise, go to the next step.

\item {({\tt S4$_{\tt new}$})} Introduce extra edges among the vertices corresponding to a subset in a nested set. The types of these edges are also identified with {({\tt S2$_{\tt new}$})}.
\end{itemize}
\vspace{-2mm}


The novelty of the new method lies in getting rid of previous two limitations by defining the types of edges and introducing extra edges.
Only the edges with the same type can be permuted with each other, which restricts the permutability of vertices. Then, the vertices in each type are permutable (i.e., these vertices can be permuted but may require other vertices to be permuted together) but may not be arbitrarily permutable (i.e., these vertices can be permuted arbitrarily no matter how other vertices are permuted).


With the new method, the graph for the policy ${\bm \rho} = g_{\tt D2D}({\bf H})$ is with two types of edges as illustrated in Fig. \ref{fig:e1}. One type of edges reflects the communication links (i.e., the solid lines), and the other reflects the interference links (i.e., the dashed lines).
The types of edges restrict the Tx vertex and Rx vertex in a transceiver pair to be re-ordered together. Then, Tx vertices or Rx vertices are permutable but are not arbitrarily permutable.
The function defined on the graph satisfies the joint PE property.

For the policy ${\bf v} = g_{\tt CoMP}({\bf h})$, according to the new method, we introduce edges among the AN vertices at each BS since the antennas at each BS constitute a subset of the nested set. Intuitively, the extra edges reflect the interaction among the AN vertices at a BS since the actions are subject to the power constraint at the BS. The edges among the AN vertices at each BS belong to one type, and the edges between the AN vertices and the UE vertex belong to another type, since these two types of edges are not permutable. The constructed graph is illustrated in Fig. \ref{fig:e2}, on which the function satisfies the nested permutation property.


\subsection{How to Apply the Proposed method}\label{section:examples}
To showcase how to construct a graph using the new method such that the function defined on the graph satisfies the desired permutation property, we consider two representative problems that consist of multiple kinds of sets. One is the multi-cell-multi-user (MCMU) power allocation, the other is hybrid analog and baseband precoding.

{\em \textbf{{Example 1 (MCMU-Power Allocation):}}}
Consider a downlink cellular system with $M$ cells that share the same spectrum, where each BS with multiple antennas serves $K$ single antenna UEs.

After beamforming, each BS transmits to each UE associated with it through an equivalent channel, which can be regarded as the channel between the UE and an equivalent antenna at the BS (called ${\rm AN}$ for short). In other words, the AN ``transmits'' to  the UE through the equivalent channel. An AN and its ``served'' UE is called as an AN-UE pair.
The equivalent channel gain between the $k$-th UE in cell$_m$ (denoted as UE$_{k_m}$) and the $j$-th ${\rm AN}$ at BS$_n$ (denoted as ${\rm AN}_{j_n}$) is $h_{k_mj_n} \triangleq \sqrt{{\bf g}_{k_m n}^{\sf H} {\bf w}_{j_n} {\bf w}_{j_n}^{\sf H} {\bf g}_{k_m n} }$, where ${\bf g}_{k_m n}$ is the channel vector from BS$_n$ to UE$_{k_m}$, and ${\bf w}_{j_n}$ is the beamforming vector from BS$_n$ to UE$_{j_n}$.

\vspace{1mm}
\noindent  {\emph{1) Problem and the Resulting Policy}}\vspace{1mm}

Denote ${\bf H} \in \mathbb{R}^{KM \times KM}$ as the equivalent channel matrix of all UEs, whose $m$-th row and $n$-th column is ${\bf H}^{mn} \in \mathbb{R}^{K \times K}$,
which is the equivalent channel matrix from the ${\rm AN}$s at BS$_n$ to the UEs in cell$_m$.


Denote the MCMU-power allocation policy as ${\bf P} = f_{\tt MCMU}({\bf H})$, where
${\bf P}={\rm diag}([{\bf p}_1, \cdots, {\bf p}_M]) \in \mathbb{R}^{KM \times KM}$, ${\bf p}_m=[p_{1_m}, \cdots, p_{K_m}] \in \mathbb{R}^{1 \times K} $, and $p_{k_m}$ is the power allocated to UE$_{k_m}$.
The powers can be optimized from a problem, say for maximizing the spectral efficiency (SE) under the power
constraint as follows,
\vspace{-2mm}
\begin{subequations}
\begin{align}
\label{eqn:problem-power}
{\rm P1}\!\!: \!\max_{{\bf P}} \!\! & \sum_{m=1}^{M}\!\sum_{k=1}^{K} \! \log_2\!\!\left(\!\!1\!+\!\frac{|h_{k_mk_m}|^2 p_{k_m} }{\sum_{(j,n)\neq (k,m)}\!|h_{k_mj_n}|^2 p_{j_n}\!\!+\!{\sigma_0^2}}\!\! \right)\!\!,\\[-3mm]
{\rm s.t.} \hspace{1mm} & 0 \leq \sum_{k=1}^{K} p_{k_m} \leq {P}^{\max},  \\[-6mm] \nonumber
\end{align}
\end{subequations}
where ${\sigma_0^2}$ is the noise power, and ${P}^{\max}$ is the maximal transmit power of each BS.

\vspace{1mm}
\noindent  {\emph{2) Sets and Permutation Property}}\vspace{1mm}

It is easy to show that the values of the objective and constraint in ${\rm P1}$ remain unchanged when the AN-UE pairs in each cell are re-ordered or the cells are re-ordered.
Hence, the policy $g_{\tt MCMU}(\cdot)$ is defined on AN set and UE set, and both are nested sets: $\{\{\text {{\rm AN}s in cell$_1$}\}, \! \cdots \!, \! \{\text{{\rm AN}s in cell$_M$} \!\}\}$, and $\{\{\text {UEs in cell$_1$}\}, \! \cdots \!, \! \{\text{UEs in cell$_M$\!} \}\}$. The two nested sets are related, since an ${\rm AN}$ and its served UE should be re-ordered as a whole.

We take the ${\rm AN}$ set as an example to analyze the permutation property of the policy.
Denote ${\bm \Pi}_{\tt c} \in \{0, 1\}^{M \times M}$ as the permutation matrix for re-ordering the indices of cells.
Since the $K$ ${\rm AN}$s in each cell are re-ordered together when the cells are re-ordered, we use ${\bm \Pi}_{\tt c} \otimes {\bf I}_{K}$ to represent the permutation of all ${\rm AN}$s in all cells when the cells are re-ordered.
Since the ${\rm AN}$s in every cell can be re-ordered, we use $M$ permutation matrices to represent the permutations. Denote ${\bm \Pi}_{{\tt AU},m} \in \{0, 1\}^{K \times K}$ as the permutation matrix that represents re-ordering the indices of ${\rm AN}$s in cell$_m$.
Then, when ${\rm AN}$s are re-ordered, ${\bf P}$ and ${\bf H}$ are respectively permuted as ${\bm \Omega}{\bf P}$ and ${\bm \Omega}{\bf H}$, where ${\bm \Omega} =({\bm \Pi}_{\tt c} \otimes {\bf I}_{K}){\rm diag}({\bm \Pi}_{{\tt AU},1}, \cdots, {\bm \Pi}_{{\tt AU},M}) \in \{0, 1\}^{KM \times KM}$.

For the UE sets, the permutation matrix multiplying with ${\bf P}$ and ${\bf H}$ is also ${\bm \Omega}$ when the UEs are re-ordered, because an ${\rm AN}$ and its served UE should be re-ordered jointly. Hence, when the ANs and UEs are re-ordered simultaneously, ${\bf P}$ and ${\bf H}$ are permuted as ${\bm \Omega} {\bf P} {\bm \Omega}^{\sf T}$ and ${\bm \Omega}{\bf H}{\bm \Omega}^{\sf T}$, respectively.

As a result, the MCMU-power allocation policy exhibits a joint and nested permutation property as follows,
\begin{equation}
{\bm \Omega} {\bf P} {\bm \Omega}^{\sf T}=f_{\tt MCMU}({\bm \Omega}{\bf H}{\bm \Omega}^{\sf T}). \label{eqn:joint-and-nested} \vspace{-1mm}
\end{equation}
%

{\bf \emph{Remark 2}:}
This problem can also be regarded as consisting of a nested set: $\{\{\text {{\rm AN}-UE pairs in cell$_1$}\}, \! \cdots \!, $ $ \! \{\text{{\rm AN}-UE pairs in cell$_M$} \}\}$, where each subset consists of the AN-UE pairs in each cell. The permutation property of the policy defined on the set is the same as  \eqref{eqn:joint-and-nested}.

\vspace{1mm}
\noindent  {\emph{3) Graph Construction}}\vspace{1mm}

\begin{figure}[!htb]
\centering
\subfigure[Graph constructed for MCMU-power allocation, $M=2$, $K=2$]{
\includegraphics[width=0.32\textwidth]{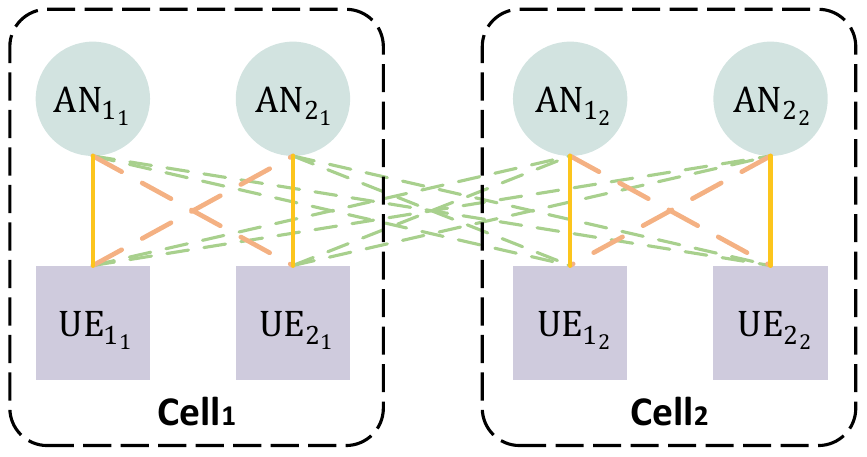}
\label{fig:example-2}
} \\
\subfigure[Graph constructed for hybrid precoding, $N_t=3$, $N_R=3$, $K=2$, $N_r=2$]{
\includegraphics[width=0.33\textwidth]{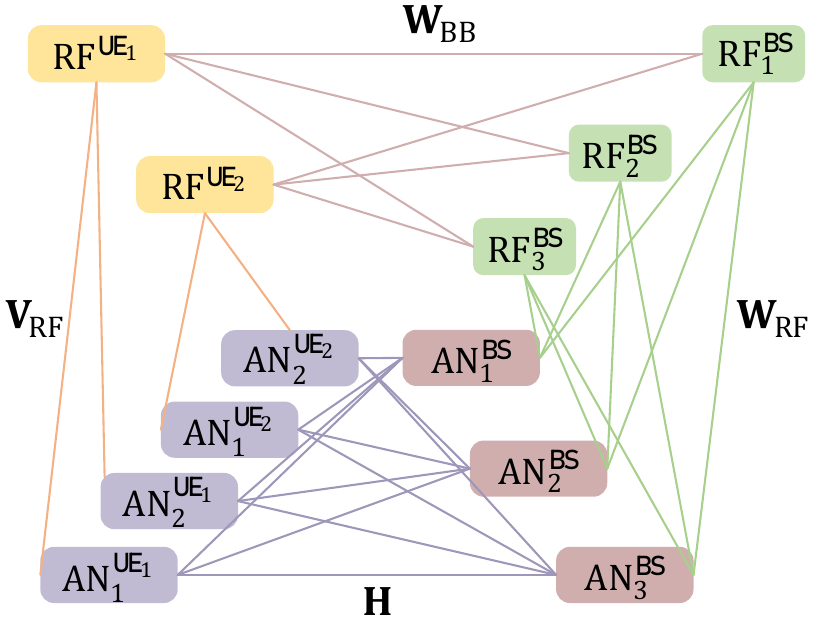}
\label{fig:example-3}
}
\caption{Graphs constructed with the proposed method, the vertices or the edges with the same color belong to the same type}
\label{fig:examples}
\vspace{2mm}
\end{figure}

We construct a graph for the problem such that  the function defined on the graph satisfies the property in \eqref{eqn:joint-and-nested}. By using the proposed method as detailed below, the constructed graph is illustrated in Fig. \ref{fig:example-2}.

According to {({\tt S1$_{\tt new}$})}, each ${\rm AN}$ is a vertex, each UE is another type of vertex.
Both  ${\rm AN}$ vertices and UE vertices have no features. The action ${\bf P}$ is defined on AN vertices.

According to {({\tt S2$_{\tt new}$})}, the edges and their features are defined from the channel matrix.
Since ${\bf H}$ is with AN dimension and UE dimension, and the action at each AN affects the received signal at each UE, we define an edge between each AN vertex and each UE vertex.
The feature of edge $(k_m, j_n)$ (i.e., the edge connecting the AN$_{k_m}$ vertex and the UE$_{j_n}$ vertex) is $|h_{k_mj_n}|^2$.

\vspace{2mm}

Next, we demonstrate how to define the types of edges. We take the system setting in Fig. \ref{fig:example-2} as an example. Since all AN-UE pairs constitute a nested set meanwhile each AN and each UE respectively correspond to a AN vertex and a UE vertex, the AN vertices and the UE vertices in each cell need to be re-ordered together. According to the permutability of the AN and UE vertices, we can identify which edges can be permuted with each other. Specifically, in cell$_1$, edge $(1_1, 1_1)$ and edge $(2_1, 2_1)$ can be permuted, and edge $(1_1, 2_1)$ and edge $(2_1, 1_1)$ are permutable, but edge $(1_1, 1_1)$ or edge $(2_1, 2_1)$ cannot be permuted with edge $(1_1, 2_1)$ or edge $(2_1, 1_1)$.
Then, edge $(1_1, 1_1)$ and edge $(2_1, 2_1)$ are with the same type, edge $(1_1, 2_1)$ and edge $(2_1, 1_1)$ are with another type.
Since edges $(1_1, 1_1)$, $(2_1, 2_1)$, $(1_2, 1_2)$ and $(2_2, 2_2)$ can be permuted with each other, hence they are with the same type (called ${\rm COM}$ edge).  Similarly, edges $(1_1, 2_1)$, $(2_1, 1_1)$, $(1_2, 2_2)$ and $(2_2, 1_2)$ belong to another type (called ${\rm MUI}$ edge). Furthermore, some edges (e.g., edge $(2_1, 1_2)$) cannot be permuted with either ${\rm COM}$ or ${\rm MUI}$ edges, which are defined as the third type of edges (called ${\rm ICI}$ edges).
The action ${\bf P}$ can also be defined on ${\rm COM}$ edges.

We can see that the types of edges constrain the permutability of vertices, i.e., the vertices in each type cannot be arbitrarily permuted.

It is not hard to show that the function defined on the graph with the three types of edges satisfies the property in \eqref{eqn:joint-and-nested}. According to {({\tt S3$_{\tt new}$})}, the graph construction is completed.

{\em \textbf{{Example 2 (Hybrid Precoding):}}}
Consider a downlink MU-MIMO mmWave communication system, where a BS equipped with $N_t$ antennas and $N_R$ radio frequency (RF) chains serves $K$ UEs, each receiving a data stream with $N_r$ antennas and a RF chain.

\vspace{1mm}
\noindent  {\emph{1) Problem and the Resulting Policy}}\vspace{1mm}

Denote ${\bf W}_{\tt RF} \in \mathbb{C}^{N_t \times N_R}$ as the analog precoder, ${\bf W}_{\tt BB} = [{\bf W}_{{\tt BB}, 1}, \cdots, {\bf W}_{{\tt BB}, K}] \in \mathbb{C}^{N_R \times K}$ as the baseband precoder, and ${\bf V}_{\tt RF} = {\rm diag}({\bf V}_{{\tt RF}, 1}, \cdots, {\bf V}_{{\tt RF}, K}) \in \mathbb{C}^{KN_r \times K}$ as the analog combiner, where ${\bf W}_{\tt BB,k}$ and  ${\bf V}_{\tt RF,k}$ are the baseband precoder and analog combiner of UE$_k$, respectively.

Denote the hybrid precoding policy as $({\bf W}_{\tt RF},{\bf W}_{\tt BB}, {\bf V}_{\tt RF})$ $=$ ${f_{\tt H}}({\bf H})$, where ${\bf H} =[{\bf H}_1^\mathsf{T}, \cdots, {\bf H}_K^\mathsf{T}]^\mathsf{T}$ $\in$ $\mathbb{C}^{KN_r \times N_t}$.

${\bf W}_{\tt RF}$, ${\bf W}_{\tt BB}$, ${\bf V}_{\tt RF}$ can be jointly optimized from a problem, say the problem that maximizes the SE as follows,
\vspace{-1mm}
\begin{subequations}
\small
\begin{align}
{\rm P2}: \max_{\substack{{\bf W}_{\tt RF},{\bf W}_{\tt BB}, \\ {\bf V}_{\tt RF}}}&\sum_{k=1}^{K}\! \log_2 \left(\! 1 \!+\! \frac{\frac{P^{\max}}{K}|{\bf V}_{{\tt RF}, k}^{\sf H}{\bf H}_k{\bf W}_{\tt RF}{\bf W}_{{\tt BB}, k}|^2}{\frac{P^{\max}}{K}\!\!\sum\limits_{n \neq k}\!\!|{\bf V}_{{\tt RF}, k}^{\sf H}{\bf H}_k\!{\bf W}_{\tt RF}\!{\bf W}_{{\tt BB}, n}|^2 \!+\! \sigma_0^2}\right)\!,  \label{eqn:obj3} \\
{\rm s.t.} \hspace{2mm} & {\rm Tr}({\bf W}_{\tt RF}{\bf W}_{\tt BB}{\bf W}_{\tt BB}^{\sf H}{\bf W}_{\tt RF}^{\sf H}) = K, \label{eqn:max-power}\\
& |{{W}_{\tt RF}}_{ij}|^2 = 1/N_t, \forall i, j, \label{eqn:ana1}\\[-1.5mm]
& |{{V}_{{\tt RF}, k}}_{ij}|^2 = 1/N_r, \forall i, j, k,  \label{eqn:ana2} \\[-7mm] \nonumber
\end{align}
\end{subequations}
where $P^{\max}$ is the maximal transmit power at the BS, ${\bf H}_k \in \mathbb{C}^{N_r \times N_t}$ is the channel matrix from  the BS to UE$_k$, and $\sigma_0^2$ is the noise power. \eqref{eqn:max-power} is the power constraint, \eqref{eqn:ana1} and \eqref{eqn:ana2} are respectively the constant modulus constraints for the analog precoder and analog combiner.

\vspace{1mm}
\noindent  {\emph{2) Sets and Permutation Property}}\vspace{1mm}

It is not hard to show that the values of the objective and constraints remain unchanged when the RF chains at the BS (RFs$^{\tt BS}$ for short) or the antennas at the BS (ANs$^{\tt BS}$ for short) are re-ordered,
the antennas at UEs (ANs$^{\tt UE}$ for short) and RF chains at UEs (RFs$^{\tt UE}$ for short)  are re-ordered together, or the antennas at each UE are re-ordered.
Hence, the policy ${f_{\tt H}}(\cdot)$ is defined on four sets as follows,
\begin{equation}
\nonumber
\begin{aligned}
&{\text {RF}^{\tt BS} ~ {\text {set}}}: &&\!\!\!\!\{{\text {RF}}_1^{\tt BS}, \cdots, {\text {RF}}_{N_R}^{\tt BS}\}, \\
&{\text {AN}}^{\tt BS} ~ {\text {set}} : &&\!\!\!\!\{{\text {AN}}_1^{\tt BS}, \cdots, {\text {AN}}_{N_t}^{\tt BS}\}, \\
&{\text {RF}}^{{\tt UE}} ~ {\text {set}}: &&\!\!\!\!\{{\text {RF}}^{{\tt UE}_1}, \cdots, {\text {RF}}^{{\tt UE}_K} \},  \\
&{\text {AN}}^{{\tt UE}} ~ {\text {set}}: &&\!\!\!\!\{\{{\text {AN}}_{1}^{{\tt UE}_1}\!, \!\cdots\!, {\text {AN}}_{N_r}^{{\tt UE}_1} \}, \! \cdots\!, \! \{{\text {AN}}_{1}^{{\tt UE}_K}\!, \cdots\!, {\text {AN}}_{N_r}^{{\tt UE}_K}\! \}\}.
\end{aligned}
\end{equation}
The RF$^{\tt BS}$ set, AN$^{\tt BS}$ set and RF$^{\tt UE}$  set are normal sets, while the AN$^{\tt UE}$ set is a nested set with $K$ subsets.
The RF$^{\tt UE}$ set and AN$^{\tt UE}$ set are related since the RF chain and the ANs at the same UE need to be re-ordered together.

When RFs$^{\tt BS}$ and ANs$^{\tt BS}$ are re-ordered, ${\bf W}_{\tt RF}$, ${\bf W}_{\tt BB}$ and ${\bf H}$ are permuted as ${\bm \Pi}_{\tt a}{\bf W}_{\tt RF}{\bm \Pi}_{\tt RF}^\mathsf{T}$, ${\bm \Pi}_{\tt RF}{\bf W}_{\tt BB}$ and ${\bf H}{\bm \Pi}_{\tt a}^\mathsf{T}$, where ${\bm \Pi}_{\tt RF} \in \{0, 1\}^{N_R \times N_R}$ and ${\bm \Pi}_{\tt a} \in \{0, 1\}^{N_t \times N_t}$ are the permutation matrices that represent re-ordering RFs$^{\tt BS}$ and ANs$^{\tt BS}$, respectively.

When the RFs$^{\tt UE}$ are re-ordered, ${\bf W}_{\tt BB}$ and ${\bf V}_{\tt RF}$ are respectively permuted as ${\bf W}_{\tt BB}{\bm \Pi}_{\tt u}^{\sf T}$ and ${\bf V}_{\tt RF}{\bm \Pi}_{\tt u}^{\sf T}$, where ${\bm \Pi}_{\tt u} \in \{0, 1\}^{K \times K}$ is the permutation matrix that represents re-ordering the RFs$^{\tt UE}$.

Considering that RF$^{\tt UE}$ set and AN$^{\tt UE}$ set are related and there are $N_r$ ANs at each UE, we use ${\bm \Pi}_{\tt u} \otimes {\bf I}_{N_r}$ to represent the permutation of all ANs at all UEs when the RFs$^{\tt UE}$ are re-ordered. Since the ANs at an UE can be re-ordered independently from the ANs at another UE, we use $K$ permutation matrices to represent the permutations, i.e., denote ${\bm \Pi}_{{\tt a},k} \in \{0, 1\}^{N_r \times N_r}$ to represent re-ordering the ANs$^{{\tt UE}_k}$. Then, when the ANs$^{\tt UE}$ are re-ordered, ${\bf V}_{\tt RF}$ and ${\bf H}$ are respectively permuted as ${\tilde {\bm \Omega}}{\bf V}_{\tt RF}$ and ${\tilde {\bm \Omega}}{\bf H}$, where ${\tilde {\bm \Omega}}$ $=({\bm \Pi}_{\tt u} \otimes {\bf I}_{N_r}){\rm diag}({\bm \Pi}_{{\tt a},1}, \cdots, {\bm \Pi}_{{\tt a},K}) \in \{0, 1\}^{KN_r \times KN_r}$.

Consequently, the hybrid precoding policy satisfies a joint, independent and nested permutation property as follows,
\begin{equation}
\label{eqn:per-precoding}
({\bm \Pi}_{\tt a}{\bf W}_{\tt RF}{\bm \Pi}_{\tt RF}^\mathsf{T}, \!{\bm \Pi}_{\tt RF}{\bf W}_{\tt BB}{\bm \Pi}_{\tt u}^\mathsf{T}, {\tilde {\bm \Omega}}{\bf V}_{\tt RF}{\bm \Pi}_{\tt u}^\mathsf{T}\!)\!=\!f_{\tt H}({\tilde {\bm \Omega}}{\bf H}{\bm \Pi}_{\tt a}^\mathsf{T}).
\end{equation}

{\bf \emph{Remark 3}:}
If we regard the RF chain and the antennas at the same UE as an element in a normal set (as the D2D example in Remark 1), then the permutations of the antennas at each UE are overlooked, and the resulting permutation property is not the same as \eqref{eqn:per-precoding}.

\vspace{1mm}
\noindent  {\emph{3) Graph Construction}}\vspace{1mm}

We construct a graph for the problem such that the function defined on the graph satisfies the property in \eqref{eqn:per-precoding}. By using the proposed method as detailed below, the constructed graph is illustrated in Fig. \ref{fig:example-3}.

According to {({\tt S1$_{\tt new}$})}, there are four types of vertices: RF$^{\tt BS}$ vertices, AN$^{\tt BS}$ vertices, AN$^{\tt UE}$ vertices and RF$^{\tt UE}$ vertices. All vertices are without features and actions.

According to {({\tt S2$_{\tt new}$})}, there is an edge between each AN$^{\tt BS}$ vertex and AN$^{\tt UE}$ vertex, since ${\bf H}$ is with AN$^{\tt BS}$ and AN$^{\tt UE}$ dimensions. Similarly, we can define the edges from the dimensions of ${\bf W}_{\tt RF}, {\bf W}_{\tt BB}, {\bf V}_{\tt RF}$. The features of the edges between the AN$^{{\tt UE}_k}$ vertices and the AN$^{\tt BS}$ vertices are ${\bf H}_k$. Other edges have no features. All actions, i.e., ${\bf W}_{\tt RF}, {\bf W}_{\tt BB}, {\bf V}_{\tt RF}$, are defined on the edges.

Next, we define the types of the edges by
taking the system setting in Fig. \ref{fig:example-3} as an example.
Since the four ANs at the two UEs constitute a nested set, the two AN vertices at each UE can be swapped but the ANs at different UEs cannot. This indicates that the two edges connecting the RF chain vertex at each UE to AN 1 vertex and to AN 2 vertex at the UE are permutable, i.e., they are with the same type.
In this way, we can define four types of edges.

It is not hard to show that the function defined on the graph satisfies the property in \eqref{eqn:per-precoding}. According to {({\tt S3$_{\tt new}$})}, the graph construction is completed.

{\bf \emph{Remark 4}:} It is noteworthy that {({\tt S4$_{\tt new}$})} is useless for the two examples. For the graph constructed for MCMU-power allocation, extra edges are unnecessary among the AN vertices at each cell, because the ${\rm MUI}$ edges and ${\rm ICI}$ edges have restricted the permutability of all the AN vertices. For the graph constructed for hybrid precoding, extra edges are not required among the AN vertices at each UE, because the edges connecting the AN$^{\tt UE}$ vertices and the RF$^{\tt UE}$ vertex at each UE have reflected the hardware constraint.

\section{Designing Architectures of HetGNNs with Desired Permutation Properties} \label{section:permutation}
In this section, we first introduce HetGNNs without any permutation property, and then provide a property of their hidden representations for satisfying the permutation property induced by graph.
To show the advantage of a proposed method for designing the architecture of HetGNN, we extend an existing method of designing PENN into the design for the HetGNN and show its limitations.
Finally, we propose a method with rigorous proof for designing a HetGNN with desired permutation  property.

\subsection{HetGNNs Without Permutation Properties} \label{section:graph-hetgnn}

HetGNNs are the DNNs that update hidden representations of vertices or edges over heterogeneous graphs by pooling the processed information of neighboring vertices and edges, and combining with hidden representations in the previous layer. The processing, pooling and combining functions for different vertices or edges can be arbitrarily selected.

HetGNNs do not naturally satisfy permutation properties. To show this, we
present the HetGNNs where the processing, pooling and combining functions differ when updating the hidden representations of different vertices or edges.\footnote{These functions are usually not vertex-specific or edge-specific. For example, the three functions are respectively the same for all vertices or all edges in  HomoGNNs.} These HetGNNs can leverage the topology information by learning over graphs, but do not exhibit any permutation property, which is explained later.

We consider two kinds of HetGNNs, which are referred to as ${\text {VertexGNN}}_{\text {NP}}$ and ${\text {EdgeGNN}}_{\text {NP}}$, where ``NP'' stands for not exhibiting any permutation property.

\subsubsection{${\text {VertexGNN}}_{\text {NP}}$} \label{section:flexible-vertex}
In ${\text {VertexGNN}}$, the hidden representation of each vertex is updated over graph.
The neighboring vertices of a vertex are the vertices connected to the vertex with an edge, and the edge is called the neighboring edge of the vertex.  Denote the edge connecting the $v$-th and $u$-th vertices as edge $(v, u)$.

For ${\text {VertexGNN}}_{\text {NP}}$, the hidden representation of the $v$-th vertex in the $l$-th layer can be obtained as follows,
\begin{equation} \label{eqn:update-vertex}
\begin{aligned}
{\bf d}_{v}^{(l)}={\rm CB}_{v}\left({\bf d}_{v}^{(l-1)}, {\rm PL}_{v}\Big(q_{v, u}({\bf d}_{u}^{(l-1)}, {\bf e}_{v}^u;
{\bm \phi}_{v, u}^{(l)}), \right. \\
\left.   u \in \mathcal{N}(v) \Big);{\bm \theta}_{v}^{(l)}\right),
\end{aligned}
\end{equation}
where ${\rm CB}_{v}(\cdot,\cdot,{\bm \theta}_{v}^{(l)})$ is the combining function for the $v$-th vertex with trainable parameters ${\bm \theta}_{v}^{(l)}$, ${\rm PL}_{v}(\cdot)$ is the pooling function for the $v$-th vertex to aggregate information from its neighboring vertices and edges,
$q_{v, u}(\cdot,\cdot;{\bm \phi}_{v, u}^{(l)})$ is the processing function
with trainable parameters ${\bm \phi}_{v, u}^{(l)}$,
${\bf e}_{v}^u$ is the feature of edge $(v, u)$, and $\mathcal{N}(v)$ is the set of neighboring vertices of the $v$-th vertex.

\subsubsection{${\text {EdgeGNN}}_{\text {NP}}$} \label{section:flexible-edge}
In ${\text {EdgeGNN}}$, the hidden representation of each edge is updated over graph. For edge $(v,u)$, its neighboring vertices are the $v$-th vertex and the $u$-th vertex, and its neighboring edges are the edges connected to the $v$-th vertex and the $u$-th vertex except edge $(v,u)$.

For ${\text {EdgeGNN}}_{\text {NP}}$, the hidden representation of edge $(v,u)$ in the $l$-th layer can be obtained as follows,
\begin{equation} \label{eqn:update-edge}
\begin{aligned}
{\bf d}_{vu}^{(l)}={\rm CB}_{vu} \bigg(\!{\bf d}_{vu}^{(l-1)},  {\rm PL}_{vu} \Big(\!q_{uv, m}({\bf v}_{v}, {\bf d}_{vm}^{(l-1)};{\bm \phi}_{uv, m}^{(l)}), \\
m \in \mathcal{N}(v) \backslash u \Big),
{\rm PL}_{vu}\Big(q_{vu, m}({\bf v}_{u}, {\bf d}_{um}^{(l-1)};
{\bm \phi}_{vu, m}^{(l)}),\\
m \in \mathcal{N}(u) \backslash v \Big);{\bm \theta}_{vu}^{(l)} \!\bigg),
\end{aligned}
\end{equation}
where ${\rm CB}_{vu}(\cdot,\cdot,{\bm \theta}_{vu}^{(l)})$ is the combining function for edge $(v,u)$ with trainable parameters ${\bm \theta}_{vu}^{(l)}$, ${\rm PL}_{vu}(\cdot)$ is the pooling function for edge $(v,u)$ to aggregate the information from its neighboring vertices and edges,
$q_{uv, m}(\cdot,\cdot;{\bm \phi}_{uv, m}^{(l)})$ and $q_{vu, m}(\cdot,\cdot;{\bm \phi}_{vu, m}^{(l)})$ are two processing functions with trainable parameters ${\bm \phi}_{uv, m}^{(l)}$ and ${\bm \phi}_{vu, m}^{(l)}$,
${\bf v}_{v}$ is the feature of the $v$-th vertex,
and $\mathcal{N}(v) \backslash u$ is the set of neighboring vertices of the $v$-th vertex except the $u$-th vertex.

{\bf \emph{Remark 5}:}
We take ${\text {VertexGNN}}_{\text {NP}}$ as an example to demonstrate that it does not exhibit any permutation property. When we re-order a vertex (say the $v$-th vertex) with permutation
 operation $\pi(\cdot)$, its index becomes $\pi(v)$.
According to \eqref{eqn:update-vertex}, its hidden representation after permutation can be expressed as follows,
\begin{equation} \label{eqn:update-vertex-re}
\begin{aligned}
{\widetilde {\bf d}}_{\pi(v)}^{(l)}={\rm CB}_{\pi(v)}\left({\widetilde {\bf d}}_{\pi(v)}^{(l-1)}, {\rm PL}_{\pi(v)}\Big(q_{\pi(v), u}({\widetilde {\bf d}}_{u}^{(l-1)}, \right. \\
{\bf e}_{\pi(v)}^u; {\bm \phi}_{\pi(v), u}^{(l)}),
\left. u \in \mathcal{N}(\pi(v)) \Big);{\bm \theta}_{\pi(v)}^{(l)} \right),
\end{aligned}
\end{equation}
where $\widetilde{\bf d}_{\pi(v)}^{(l)}$ denotes the hidden representation of the $\pi(v)$-th vertex in the $l$-th layer after permutation.

Since the processing, pooling and combining functions differ for different vertices, $q_{\pi(v), u}(\cdot,\cdot; {\bm \phi}_{\pi(v), u}^{(l)})$, ${\rm PL}_{\pi(v)}(\cdot)$ and ${\rm CB}_{\pi(v)}(\cdot;{\bm \theta}_{\pi(v)}^{(l)})$ are different from $q_{v, u}(\cdot,\cdot; {\bm \phi}_{v, u}^{(l)})$, ${\rm PL}_{v}(\cdot)$ and ${\rm CB}_{v}(\cdot;{\bm \theta}_{v}^{(l)})$. Due to this, from \eqref{eqn:update-vertex} and \eqref{eqn:update-vertex-re},  it is not hard to see that ${\bf d}_{v}^{(l)} \neq {\widetilde {\bf d}}_{\pi(v)}^{(l)}$, which indicates that ${\text {VertexGNN}}_{\text {NP}}$ does not exhibit any permutation property.

Similarly, we can demonstrate that ${\text {EdgeGNN}}_{\text {NP}}$ also does not exhibit any permutation property.

\subsection{Property of Hidden Representations in HetGNNs}
Since the permutation property of a function on a graph is induced by the types of vertices and edges, the input and output of a HetGNN should be permuted accordingly when the vertices or edges with the same type in the graph are re-ordered (not necessarily arbitrarily). Because permutation property can be preserved by stacking multiple layers \cite{zaheer2017deep},
every layer of a HetGNN should satisfy the property. This indicates that the hidden representations should satisfy the following property.

%

\vspace{1mm}
{\bf Property 1:} \emph{When the indices of the vertices (or edges) with the same type in a graph are permuted (not necessarily arbitrarily), the hidden representations of these vertices (or edges) learned by a VertexGNN (or an EdgeGNN) over the graph should be permuted accordingly.}\vspace{1mm}

A HetGNN satisfying Property 1 exhibits the permutation property of a function defined on a graph.

\subsection{Extending an Existing Method to Satisfy Property 1} \label{design-GNN}
We start by briefly reviewing the method proposed in \cite{zaheer2017deep, SEquivariance,maron2019universality,hartford2018deep}, which designs a PENN to satisfy a given permutation property by introducing parameter sharing into a FNN.

\subsubsection{Recap the Method in \cite{zaheer2017deep, SEquivariance,maron2019universality,hartford2018deep} for Designing Parameter Sharing}
\label{review_penn}

This method decides how to share weights in each layer of a FNN by first enumerating all possible permutation matrices, then enumerating all sharable parameters determined by each permutation matrix.

The input-output relation of the $l$-th layer of a FNN can be expressed in matrix form as
\begin{equation}
\label{eqn:fnn}
{\bf y}^{(l)}=\sigma({\bf W}^{(l)}{\bf y}^{(l-1)}),
\end{equation}
where ${\bf y}^{(l)}$ and ${\bf y}^{(l-1)}$ are respectively the outputs of the $l$-th layer and the $(l-1)$-th layer, $\sigma(\cdot)$ is the activation function, ${\bf W}^{(l)}$ is the weight matrix composed of trainable parameters.

To help understand, we take the power allocation policy ${\bf p}=g_{\tt MU}({\bf h})$ introduced in subsection \ref{relation-ppsp} as an example, which satisfies ${\bf \Pi} {\bf p}=g_{\tt MU}({\bf \Pi} {\bf h}), \forall {\bf \Pi} \in \mathcal{P}$,
where $\mathcal{P}$ is a set of possible permutation matrices.\footnote{When $K=3$, there are $3!=6$ possible permutation matrices as follows,
\vspace{-1mm}
\begin{equation} \label{eqn:per-matrix-d2d}
\scriptsize
\begin{aligned}
{\bm \Pi}_1\!\!=\!\!
\begin{bmatrix}
0 & 1 & 0 \\
1 & 0 & 0 \\
0 & 0 & 1 \\
\end{bmatrix},
{\bm \Pi}_2\!\!=\!\!
\begin{bmatrix}
0 & 0 & 1 \\
0 & 1 & 0 \\
1 & 0 & 0 \\
\end{bmatrix},
{\bm \Pi}_3\!\!=\!\!
\begin{bmatrix}
1 & 0 & 0 \\
0 & 0 & 1 \\
0 & 1 & 0 \\
\end{bmatrix}, \\
{\bm \Pi}_4\!\!=\!\!
\begin{bmatrix}
0 & 1 & 0 \\
0 & 0 & 1 \\
1 & 0 & 0 \\
\end{bmatrix},
{\bm \Pi}_5\!\!=\!\!
\begin{bmatrix}
0 & 0 & 1 \\
1 & 0 & 0 \\
0 & 1 & 0 \\
\end{bmatrix},
{\bm \Pi}_6\!\!=\!\!
\begin{bmatrix}
1 & 0 & 0 \\
0 & 1 & 0 \\
0 & 0 & 1 \\
\end{bmatrix}.
\end{aligned} \vspace{-1mm}\nonumber
\end{equation}}

To satisfy the 1D-PE property, the input-output relation in \eqref{eqn:fnn} should satisfy ${\bf \Pi}{\bf y}^{(l)}=\sigma({\bf W}^{(l)}{\bf \Pi} {\bf y}^{(l-1)})$, $\forall {\bf \Pi} \in \mathcal{P}$. This can be accomplished by introducing parameter sharing into ${\bf W}^{(l)}$ with the method in \cite{zaheer2017deep, SEquivariance,maron2019universality,hartford2018deep}.

Specifically, for a given ${\bf \Pi} \in \mathcal{P}$, if the $n$-th element in ${\bf y}^{(l-1)}$ is permuted to the $n'$-th element in ${\bf \Pi}{\bf y}^{(l-1)}$, and the $m$-th element in ${\bf y}^{(l)}$ is permuted to the $m'$-th element in ${\bf \Pi}{\bf y}^{(l)}$, then ${W}^{(l)}_{mn}$ should be the same as $W^{(l)}_{m'n'}$.
For example, for ${\bf \Pi}_1 \in \mathcal{P}$,
the first element in ${\bf y}^{(l-1)}$ is permuted to the second element in ${\bf \Pi}_1{\bf y}^{(l-1)}$ and  the first element in ${\bf y}^{(l)}$ is permuted to the second element in ${\bf \Pi}_1{\bf y}^{(l)}$. According to the method in \cite{zaheer2017deep, SEquivariance,maron2019universality,hartford2018deep}, we have ${W}^{(l)}_{11}={W}^{(l)}_{22}$.
After further enumerating the other parameters in ${\bf W}^{(l)}$ (i.e., ${W}^{(l)}_{12}$, ${W}^{(l)}_{13}$, ${W}^{(l)}_{21}$, ${W}^{(l)}_{22}$, ${W}^{(l)}_{23}$, ${W}^{(l)}_{31}$, ${W}^{(l)}_{32}$), the parameter sharing for  ${\bf \Pi}_1$ can be designed.

The parameter sharing of ${\bf W}^{(l)}$ for other permutation matrices in $\mathcal{P}$ can be obtained similarly. After enumerating all possible permutation matrices in $\mathcal{P}$, the parameter sharing scheme is obtained, where ${\bf W}^{(l)}$ is with the following structure: the diagonal elements are identical and the off-diagonal elements are identical.

This method of designing parameter sharing is rather involved.
To see this, again consider the power allocation policy ${\bf p}=g_{\tt MU}({\bf h})$, which is defined on a single set. For a system with $K$ UEs, there are $K!$ permutation matrices in $\mathcal{P}$. After regular but tedious manipulations, the total times of enumerations required to design parameter sharing for ${\bf W}^{(l)} \in \mathbb{R}^{K \times K}$ can be derived as $N_{\rm enumer}= \sum_{n=2}^{K}[n!C_K^n - (n-1)!C_K^{n-1}] \times (2Kn-n^2)$, where $C_i^j=\frac{i!}{j!(i-j)!}$. When $K=2$, $N_{\rm enumer}=4$, when $K=3$, $N_{\rm enumer}=24$, and when $K=4$, $N_{\rm enumer}=276$.
To design the parameter sharing for a PENN to satisfy a desired permutation property, one can first design for the smallest problem sizes (say $K=2$), until the structure of the weight matrix ${\bf W}^{(l)}$ no longer changes with the number.

For the problems with multiple sets, this method will become even more tedious.

\subsubsection{Extending the Method in \cite{zaheer2017deep, SEquivariance,maron2019universality,hartford2018deep} to Design HetGNNs}
\label{section: extending}
To extend the method for introducing parameter sharing into FNN to the design of a HetGNN, one needs to express the update equation of the HetGNN in matrix form as in \eqref{eqn:fnn}.
Due to the flexible choices of processing, combining and pooling functions, the update equations of many GNNs cannot be expressed in matrix form, say the HetGNNs with non-linear processors.
To express the update equation in matrix form, the following assumption is required.

\vspace{1mm}{\bf Assumption 1:} \emph{The processing function is linear, the combining function is a one-layer FNN, and the pooling function is summation or mean function.} \vspace{1mm}

For the HetGNNs under Assumption 1, the reviewed method can be applied to design the HetGNN to satisfy Property 1 by regarding the hidden representations in the $l$-th layer and the $(l-1)$-th layer of the HetGNN as ${\bf y}^{(l)}$ and ${\bf y}^{(l-1)}$ of a FNN.



Next, we show how to use the extended method to design HetGNNs with the two example problems in section \ref{section:examples}.

\vspace{1mm}{\em \textbf{{ Example 1 (MCMU-Power Allocation):}}}
Since the actions can be defined on UE vertices or ${\rm COM}$ edges, either VertexGNN or EdgeGNN can be used to learn the MCMU-power allocation policy.
We take the ${\text {VertexGNN}}_{\text {NP}}$ in subsection \ref{section:flexible-vertex} as an example to elaborate on how to design the HetGNN to satisfy the property in \eqref{eqn:joint-and-nested}. We consider two cells and two ${\rm AN}$-UE pairs in each cell as an example, and the constructed graph is in Fig. \ref{fig:example-2}.\footnote{When $M=K=2$, ${\bm \Pi}_{\tt c}, {\bm \Pi}_{{\tt AU},1}, {\bm \Pi}_{{\tt AU},2} \in \{$
{\scriptsize $\def\arraystretch{0.7}\begin{bmatrix}
1 & 0 \\
0 & 1
\end{bmatrix},
\begin{bmatrix}
0 & 1 \\
1 & 0
\end{bmatrix}$} $
\}$, where each permutation matrix represents a possible permutation on the indices of cells or ${\rm AN}$-UE pairs in a cell. Then, according to the definition of ${\bm \Omega}$, it has eight possible values  as follows, each represents a possible permutation on the indices of four ${\rm AN}$-UE pairs,
\vspace{-1mm}
\begin{equation}
\def\arraystretch{0.7}
\scriptsize
\begin{aligned}
&{\bm \Omega}_1\!\!=\!\!
\begin{bmatrix}
0 & 1 & 0 & 0 \\
1 & 0 & 0 & 0 \\
0 & 0 & 1 & 0 \\
0 & 0 & 0 & 1 \\
\end{bmatrix},
{\bm \Omega}_2\!\!=\!\!
\begin{bmatrix}
1 & 0 & 0 & 0 \\
0 & 1 & 0 & 0 \\
0 & 0 & 0 & 1 \\
0 & 0 & 1 & 0 \\
\end{bmatrix},
{\bm \Omega}_3\!\!=\!\!
\begin{bmatrix}
1 & 0 & 0 & 0 \\
0 & 1 & 0 & 0 \\
0 & 0 & 1 & 0 \\
0 & 0 & 0 & 1 \\
\end{bmatrix}, \\
&{\bm \Omega}_4\!\!=\!\!
\begin{bmatrix}
0 & 1 & 0 & 0 \\
1 & 0 & 0 & 0 \\
0 & 0 & 0 & 1 \\
0 & 0 & 1 & 0 \\
\end{bmatrix},
{\bm \Omega}_5\!\!=\!\!
\begin{bmatrix}
0 & 0 & 1 & 0 \\
0 & 0 & 0 & 1 \\
1 & 0 & 0 & 0 \\
0 & 1 & 0 & 0 \\
\end{bmatrix},
{\bm \Omega}_6\!\!=\!\!
\begin{bmatrix}
0 & 0 & 0 & 1 \\
0 & 0 & 1 & 0 \\
1 & 0 & 0 & 0 \\
0 & 1 & 0 & 0 \\
\end{bmatrix}, \\
&{\bm \Omega}_7\!\!=\!\!
\begin{bmatrix}
0 & 0 & 1 & 0 \\
0 & 0 & 0 & 1 \\
0 & 1 & 0 & 0 \\
1 & 0 & 0 & 0 \\
\end{bmatrix},
{\bm \Omega}_8\!\!=\!\!
\begin{bmatrix}
0 & 0 & 0 & 1 \\
0 & 0 & 1 & 0 \\
0 & 1 & 0 & 0 \\
1 & 0 & 0 & 0 \\
\end{bmatrix}.
\end{aligned}\nonumber
\end{equation}}

To rewrite the update equation in  matrix form, we use mean function as pooling function in ${\text {VertexGNN}}_{\text {NP}}$ to satisfy Assumption 1.
For easy understanding, we replace ${\bf d}_{v}^{(l)}$ in \eqref{eqn:update-vertex} to ${\bf b}_{j_n}^{(l)}$ and ${\bf u}_{k_m}^{(l)}$, and replace ${\bf e}_{v}^u$ in \eqref{eqn:update-vertex} to $|h_{k_mj_n}|^2$, where ${\bf b}_{j_n}^{(l)}$ and ${\bf u}_{k_m}^{(l)}$ are the hidden representations of the AN$_{j_n}$ vertex and the UE$_{k_m}$ vertex in the $l$-th layer.
By stacking the hidden representations of all vertices in the $l$-th and the $(l-1)$-th layers into vectors, the update equation of the GNN learning over the graph in Fig. \ref{fig:example-2} can be expressed as \eqref{eqn:updating-D2D} (see next page).
\begin{figure*}
\small
\begin{equation}
\begin{aligned}
\label{eqn:updating-D2D}
\def\arraystretch{1.1}
\begin{bmatrix}
{\bf b}_{1_1}^{(l)} \\
{\bf b}_{1_2}^{(l)} \\
{\bf b}_{2_1}^{(l)} \\
{\bf b}_{2_2}^{(l)} \\
{\bf u}_{1_1}^{(l)} \\
{\bf u}_{1_2}^{(l)} \\
{\bf u}_{2_1}^{(l)} \\
{\bf u}_{2_2}^{(l)}
\end{bmatrix}
 \!=\!
\def\arraystretch{1.2}
\setlength{\arraycolsep}{2.5pt}
& \begin{bmatrix}
{\bf S}_1 & {\bf 0} & {\bf 0} & {\bf 0} & {\bf 0} &  {\bf 0} &  {\bf 0} &  {\bf 0}\\
{\bf 0} & {\bf S}_2 & {\bf 0} & {\bf 0} & {\bf 0} &  {\bf 0} &  {\bf 0} &  {\bf 0}\\
{\bf 0} &  {\bf 0}& {\bf S}_3 & {\bf 0} & {\bf 0} &  {\bf 0} &  {\bf 0} &  {\bf 0}\\
{\bf 0} &  {\bf 0}& {\bf 0} & {\bf S}_4 & {\bf 0} &  {\bf 0} &  {\bf 0} &  {\bf 0}\\
{\bf 0} & {\bf 0} &  {\bf 0} &  {\bf 0} & {\bf Y}_1 & {\bf 0} & {\bf 0} &  {\bf 0}\\
{\bf 0} & {\bf 0} &  {\bf 0}& {\bf 0} &  {\bf 0} & {\bf Y}_2  & {\bf 0} &  {\bf 0} \\
{\bf 0} & {\bf 0} &  {\bf 0} &  {\bf 0} &  {\bf 0}  & {\bf 0} & {\bf Y}_3 &  {\bf 0} \\
{\bf 0} & {\bf 0} &  {\bf 0} &  {\bf 0} &  {\bf 0}  & {\bf 0} & {\bf 0} &  {\bf Y}_4
\end{bmatrix}\!\!
\begin{bmatrix}
{\bf b}_{1_1}^{(l-1)} \\
{\bf b}_{1_2}^{(l-1)} \\
{\bf b}_{2_1}^{(l-1)} \\
{\bf b}_{2_2}^{(l-1)} \\
{\bf u}_{1_1}^{(l-1)} \\
{\bf u}_{1_2}^{(l-1)} \\
{\bf u}_{2_1}^{(l-1)} \\
{\bf u}_{2_2}^{(l-1)}
\end{bmatrix} \!\!+ \frac{1}{4}
\begin{bmatrix}
{\bf 0} & {\bf 0} & {\bf 0} & {\bf 0} & {\bf C}_{11} & {\bf C}_{12} & {\bf C}_{13} & {\bf C}_{14}\\
{\bf 0} & {\bf 0} & {\bf 0} & {\bf 0} & {\bf C}_{21} & {\bf C}_{22} & {\bf C}_{23} & {\bf C}_{24}\\
{\bf 0} & {\bf 0} & {\bf 0} & {\bf 0}&  {\bf C}_{31} & {\bf C}_{32} & {\bf C}_{33} & {\bf C}_{34}\\
{\bf 0} & {\bf 0} & {\bf 0} & {\bf 0}&  {\bf C}_{41} & {\bf C}_{42} & {\bf C}_{43} & {\bf C}_{44} \\
{\bf D}_{11} & {\bf D}_{12} & {\bf D}_{13}& {\bf D}_{14} &  {\bf 0} & {\bf 0} & {\bf 0} & {\bf 0}\\
{\bf D}_{21} & {\bf D}_{22} & {\bf D}_{23} & {\bf D}_{24}& {\bf 0} & {\bf 0} & {\bf 0} & {\bf 0}\\
{\bf D}_{31} & {\bf D}_{32} & {\bf D}_{33}& {\bf D}_{34} & {\bf 0} & {\bf 0} & {\bf 0}& {\bf 0} \\
{\bf D}_{41} & {\bf D}_{42} & {\bf D}_{43} & {\bf D}_{44} & {\bf 0} & {\bf 0} & {\bf 0}& {\bf 0} \\
\end{bmatrix}\!\!
\begin{bmatrix}
{\bf b}_{1_1}^{(l-1)} \\
{\bf b}_{1_2}^{(l-1)} \\
{\bf b}_{2_1}^{(l-1)} \\
{\bf b}_{2_2}^{(l-1)} \\
{\bf u}_{1_1}^{(l-1)} \\
{\bf u}_{1_2}^{(l-1)} \\
{\bf u}_{2_1}^{(l-1)} \\
{\bf u}_{2_2}^{(l-1)}
\end{bmatrix}
 \\
 + \frac{1}{4}
\addtocounter{MaxMatrixCols}{20}
& \begin{bmatrix}
{\bf P}_{11} & {\bf 0} & {\bf 0} & {\bf 0}  & {\bf P}_{12} & {\bf 0} & {\bf 0} & {\bf 0} & {\bf P}_{13} & {\bf 0} & {\bf 0} & {\bf 0} & {\bf P}_{14} & {\bf 0} & {\bf 0} & {\bf 0} \\
{\bf 0} &  {\bf 0} & {\bf P}_{21} & {\bf 0}  & {\bf 0} &   {\bf 0} & {\bf P}_{22} & {\bf 0} &  {\bf 0} &  {\bf 0} & {\bf P}_{23} & {\bf 0} &  {\bf 0} &  {\bf 0} & {\bf P}_{24} & {\bf 0} \\
{\bf 0} & {\bf P}_{31} & {\bf 0} & {\bf 0}  & {\bf 0} & {\bf P}_{32} &  {\bf 0} & {\bf 0} &  {\bf 0} & {\bf P}_{33} & {\bf 0} & {\bf 0} &  {\bf 0} & {\bf P}_{34} & {\bf 0} & {\bf 0} \\
{\bf 0} &  {\bf 0}  & {\bf 0}  & {\bf P}_{41} & {\bf 0} &   {\bf 0}  & {\bf 0} & {\bf P}_{42}  &  {\bf 0} &  {\bf 0} &  {\bf 0} &  {\bf P}_{43} & {\bf 0} &  {\bf 0} &  {\bf 0} &  {\bf P}_{44} \\
{\bf Q}_{11} & {\bf Q}_{12} & {\bf Q}_{13} & {\bf Q}_{14} & {\bf 0} & {\bf 0} & {\bf 0} & {\bf 0} & {\bf 0} & {\bf 0} & {\bf 0} & {\bf 0} & {\bf 0} & {\bf 0} & {\bf 0} & {\bf 0} \\
{\bf 0} & {\bf 0} & {\bf 0} & {\bf 0} & {\bf 0} & {\bf 0} & {\bf 0} & {\bf 0} & {\bf Q}_{21} & {\bf Q}_{22} & {\bf Q}_{23} & {\bf Q}_{24} & {\bf 0} & {\bf 0} & {\bf 0} & {\bf 0} \\
{\bf 0} & {\bf 0} & {\bf 0} & {\bf 0} & {\bf Q}_{31} & {\bf Q}_{32} & {\bf Q}_{33} & {\bf Q}_{34} & {\bf 0} & {\bf 0} & {\bf 0} & {\bf 0} & {\bf 0} & {\bf 0} & {\bf 0} & {\bf 0} \\
{\bf 0} & {\bf 0} & {\bf 0} & {\bf 0} & {\bf 0} & {\bf 0} & {\bf 0} & {\bf 0} & {\bf 0} & {\bf 0} & {\bf 0} & {\bf 0} & {\bf Q}_{41} & {\bf Q}_{42} & {\bf Q}_{43} & {\bf Q}_{44}
\end{bmatrix}
\begin{bmatrix}
|h_{1_1 1_1}|^2 \\
\vdots \\
|h_{1_1 2_2}|^2 \\
|h_{2_1 1_1}|^2 \\
\vdots \\
|h_{2_2 2_2}|^2 \\
\end{bmatrix}.
\end{aligned}
\end{equation}
\vspace{-6mm}
\end{figure*}
In \eqref{eqn:updating-D2D}, ${\bf S}_k, {\bf Y}_k, {\bf C}_{km}, {\bf D}_{km}, {\bf P}_{km}, {\bf Q}_{km}$ are sub-matrices with trainable parameters, which are \emph{without any parameter sharing}. This validates that ${\text {VertexGNN}}_{\text {NP}}$ does not have permutation property. The zeros (i.e., ``{\bf 0}''s) come from \emph{harnessing the topology information} by passing messages over the graph. The superscript $(l)$ in the sub-matrices and the activation function in combination function are omitted for notational simplicity.


To design the parameter sharing for the ${\text {VertexGNN}}_{\text {NP}}$ to satisfy Property 1, we can enumerate all sharable parameters determined by each possible permutation of ${\bm \Omega}$, using the method reviewed in subsection \ref{review_penn}.

For instance, ${\bm \Omega}_1$ represents that the indices of the UE$_{1_1}$ and UE$_{2_1}$ vertices are swapped, meanwhile the indices of the AN$_{1_1}$ and AN$_{2_1}$ vertices are swapped, which lead to the permutation of
the features of the neighboring edges of the four vertices.
To satisfy Property 1, the hidden representations of the UE$_{1_1}$ and UE$_{2_1}$ vertices (i.e., ${\bf u}_{1_1}^{(l)}$ and ${\bf u}_{2_1}^{(l)}$, $\forall l$) should be swapped, and the hidden representations of the AN$_{1_1}$ and AN$_{2_1}$ vertices (i.e., ${\bf b}_{1_1}^{(l)}$ and ${\bf b}_{2_1}^{(l)}$, $\forall l$) should be swapped.
Same as the design of PENN, this can be accomplished by sharing the sub-matrices in ${\bf M}_1$, ${\bf M}_2$ and ${\bf M}_3$ in \eqref{eqn:updating-D2D} as ${\bf S}_{1}={\bf S}_{3}$, ${\bf C}_{11}={\bf C}_{33}$, ${\bf C}_{12}={\bf C}_{32}$, ${\bf C}_{13}={\bf C}_{31}$, ${\bf C}_{14}={\bf C}_{34}$, ${\bf D}_{11}={\bf D}_{22}$, ${\bf D}_{12}={\bf D}_{21}$, ${\bf D}_{13}={\bf D}_{31}$, ${\bf D}_{14}={\bf D}_{34}$, ${\bf P}_{11}={\bf P}_{32}$, ${\bf P}_{12}={\bf P}_{31}$, ${\bf P}_{13}={\bf P}_{33}$, ${\bf P}_{14}={\bf P}_{34}$, ${\bf Q}_{11}={\bf Q}_{32}$, ${\bf Q}_{12}={\bf Q}_{31}$, ${\bf Q}_{13}={\bf Q}_{33}$, ${\bf Q}_{14}={\bf Q}_{34}$.
The corresponding parameter sharing for the permutation matrices ${\bm \Omega}_2$ $\sim$ ${\bm \Omega}_8$ can be obtained similarly.

Analogously, the parameter sharing for the ${\text {EdgeGNN}}_{\text {NP}}$ to satisfy the joint and nested PE property can be designed, which is not provided to avoid redundancy.

For $M$ cells each with $K$ AN-UE pairs, the number of all possible permutation matrices for permuting the indices of cells and AN-UE pairs is $M!\times (K!)^M$. For each permutation matrix, we need to enumerate all the sharable parameters determined by the permutation matrix to design the parameter sharing for HetGNN. The times of enumerating sharable parameters are far larger than the number of all possible permutation matrices.


\vspace{1mm}{\em \textbf{{ Example 2 (Hybrid Precoding):}}}
Since all actions are defined on edges, using EdgeGNN is simpler, otherwise a readout layer is required to convert the hidden representations of vertices into the actions.

To introduce parameter sharing into ${\text {EdgeGNN}}_{\text {NP}}$ in subsection \ref{section:flexible-edge} to satisfy Property 1, we need to enumerate all sharable parameters
determined by each permutation matrix, using the method in \cite{zaheer2017deep, SEquivariance,maron2019universality,hartford2018deep}. For this problem, the number of all possible permutation matrices is $K!\times (N_r!)^K \times N_R! \times N_t!$, where $K!$ is the number of possible permutations of the indices of UEs, $N_r!$ is the number of possible permutations of the indices of ANs at each UE, $N_R!$ and $N_t!$ are respectively the numbers of possible permutations of the indices of ANs and RF chains at the BS. Since the number of sharable parameters is much more than the number of possible permutation matrices, designing parameter sharing for this policy is very hard.

\subsubsection{Limitation of the Extended Method}
The extended method of designing parameter sharing has the following two limitations.

(1) The method is only applicable to the HetGNNs that satisfy Assumption 1.

(2) The procedure of designing parameter sharing is tedious, especially when there are multiple types of vertices and edges in a graph.

\subsection{A New method for Designing HetGNNs with Property 1} \label{section:simple}
We begin with proving three sufficient conditions for VertexGNNs satisfying Property 1, from which
a method is proposed for designing VertexGNNs to satisfy the permutation
property of a function defined on a graph.
Then, we extend the conditions and the method to the HetGNNs that update the hidden representation of each edge and update the hidden representations of both vertices and edges.

\begin{proposition} \label{pro1}
\emph{A VertexGNN will satisfy Property 1 if the three functions of the update equation in \eqref{eqn:update-vertex} satisfy the following conditions: (C1) $q_{v_1, n_1}(\cdot)$ and $q_{v_2, n_2}(\cdot)$ are identical if (i) the $v_1$-th and the $v_2$-th vertices are with the same type, (ii) the $n_1$-th and  the $n_2$-th vertices are with the same type, and (iii) edge $(v_1, n_1)$ and edge $(v_2, n_2)$ are with the same type, (C2) ${\rm CB}_{v_1}(\cdot)$ and ${\rm CB}_{v_2}(\cdot)$ are identical, ${\rm PL}_{v_1}(\cdot)$  and ${\rm PL}_{v_2}(\cdot)$ are identical if the $v_1$-th and the $v_2$-th vertices are with the same type, and (C3)  ${\rm PL}_{v}(\cdot)~\forall v$ satisfies commutative law.}
\end{proposition}
\emph{Proof:} See Appendix A.
\vspace{1mm}

Condition (C1) indicates that the processors should be shared if the type of the vertices aggregating information, the type of the vertices to be aggregated, and the type of the neighboring edges are all the same.

Condition (C2) means that the combining and pooling functions should be shared for the vertices with same type.

Condition (C3) indicates that all pooling functions should satisfy the commutative
law, which is common for all GNNs, including HomoGNNs.

\vspace{1mm}{\em \textbf{{Proposed method:}}} The proposition suggests a method: {{\emph {share the processing, combining, and pooling functions according to the types of vertices and edges}}}.

With such \emph{function sharing}, we no longer need to enumerate all possible permutations, and no longer need to express the update equation of a HetGNN in matrix form such that Assumption 1 is unnecessary any more.
If the processing, combining and pooling functions satisfy Assumption 1, the HetGNN designed by the function sharing method is with the same parameter sharing designed by the extended method.

Next, we show how to apply the method to design HetGNNs with previous two examples.

\vspace{1mm}{\em \textbf{{ Example 1 (MCMU-Power Allocation):}}}
There are two types of vertices (${\rm AN}$ and UE vertices) and three types of edges (${\rm COM}$, ${\rm MUI}$ and ${\rm ICI}$ edges) in the constructed graph, as illustrated  in Fig. \ref{fig:example-2}.

To satisfy Condition (C1) in Proposition \ref{pro1}, six processors are required for aggregation, which can be selected as any functions such as FNNs. To satisfy Conditions (C2) and (C3), all ${\rm AN}$ vertices or all UE vertices should use the same pooling function and combining function. When the two functions are respectively selected as a mean function and a one-layer FNN, the update equation of the designed HetGNN can be expressed in matrix form with the same parameter sharing structure as in subsection \ref{section: extending}.

\vspace{1mm}{\em \textbf{{Example 2 (Hybrid Precoding):}}}
There are four types of vertices and four types of edges  in the constructed graph, as illustrated in Fig. \ref{fig:example-3}.

To satisfy Condition (C1), eight processors are required for aggregation, which can be selected as any functions.
To satisfy Conditions (C2) and (C3), all vertices with the same type should use the same pooling and combining functions.

The sufficient conditions for EdgeGNNs satisfying Property 1 are proved as follows, where proof is similar to that for Proposition \ref{pro1} and hence is omitted.

\begin{proposition} \label{pro2}
\emph{An EdgeGNN will satisfy Property 1 if the three functions of the update equation in \eqref{eqn:update-edge} satisfy the following conditions: (C1) $q_{n_1v_1,m_1}(\cdot)$ and $q_{n_2v_2,m_2}(\cdot)$ are identical if (i) edge $(n_1, v_1)$ and edge $(n_2, v_2)$ are with the same type, (ii) the $v_1$-th and  the $v_2$-th vertices are with the same type, and (iii) edge $(v_1, m_1)$ and edge $(v_2, m_2)$ are with the same type, (C2) ${\rm CB}_{v_1n_1}(\cdot)$ and ${\rm CB}_{v_2n_2}(\cdot)$ are identical, ${\rm PL}_{v_1n_1}(\cdot)$ and ${\rm PL}_{v_2n_2}(\cdot)$ are identical if edge $(n_1, v_1)$ and edge $(n_2, v_2)$ are with the same type, and (C3) ${\rm PL}_{vn}(\cdot)~\forall v, n$ satisfies commutative law.}
\end{proposition}

The proposition suggests a similar method for designing EdgeGNNs to satisfy Property 1 by sharing the three functions according to the types of vertices and edges.

\vspace{1mm}{\bf \emph{Remark 6}:}  Since the types of vertices and edges are much fewer than all possible permutations matrices and are further with dramatically less numbers than the times of enumerations of all sharable parameters, designing HetGNNs with function sharing is much easier to implement than the extended method with parameter sharing.

\vspace{1mm}{\bf \emph{Remark 7}:}
The proposed method can be extended to design other kinds of HetGNNs, such as graph attention networks \cite{velivckovic2018graph}, multidimensional GNNs (MDGNN) \cite{Multidimensional2023Liu} and the HetGNNs updating the representations of vertices and edges simultaneously \cite{battaglia2018relational,zhaoris2023}. For example, Proposition \ref{pro2} can be used to design the MDGNN for satisfying desired permutation property simply by treating the hyper-edges as edges. For designing the HetGNNs that update vertices and edges simultaneously, Proposition \ref{pro1} and Proposition \ref{pro2} can be respectively used for designing the updating equations for vertices and edges.

\section{Systematic Design for Exploiting Permutation Prior}
\label{section:V}
The heterogeneous graph and HetGNN can be designed systematically to exploit permutation prior for a wireless problem, by using the proposed methods as follows.

\vspace{-1mm}
\begin{itemize}
\item {\em Step 1:} Identify the sets of the problem.
\item {\em Step 2:} Construct a graph according to the sets using the method (i.e., {({\tt S1$_{\tt new}$})}-({\tt S4$_{\tt new}$}))  in Section \ref{section:graph-gnn}.
\item {\em Step 3:} Design the architecture of the HetGNN according to the constructed graph using the methods with function sharing in Section \ref{section:permutation}.
\end{itemize}
\vspace{-1mm}

In Table \ref{table: comparison} (see next page), we compare the proposed methods with the existing methods.
\setlength{\tabcolsep}{2pt}
\renewcommand\arraystretch{1}
\begin{table*}[htb!]
\centering
\footnotesize
\caption{Comparison With Existing Methods} \label{table: comparison}
\begin{tabular}{c|c|c}
\hline\hline
\multirow{7}{*}{Design graphs} &  \cite{Eisen2020,shen2021graph,guo2022learning, zhao2022understanding, Guo2023Generalizable, Wang2023Learning, Jiang2021graph,Chowdhury2021Unfolding,Zhang2023GNN,Chen2022graph, He2022graphtcom, Yang2024Graph,Shelim2023Learning,MishraGNN2024} &  Heuristically design graphs \\ \cline{2-3}
                  &  \cite{Multidimensional2023Liu} &  \begin{tabular}[c]{@{}c@{}}Systemically design graphs, but the types of edges are not defined \\ and the relation across multiples sets is neglected, which cannot \\ ensure HetGNNs to satisfy some complicated permutation properties \end{tabular} \\ \cline{2-3}
                  &  Proposed & \begin{tabular}[c]{@{}c@{}}Systemically design graphs, the types of edges are defined and \\ the relation across multiples sets is considered by defining edges, \\ which can ensure HetGNNs to satisfy complicated permutation properties \end{tabular} \\ \hline
\multirow{5}{*}{Design architectures of GNNs} &  \cite{Chowdhury2021Unfolding, shen2021graph, Zhang2023GNN,Shelim2023Learning, Zhangmodelgnn2025} &  Design homogeneous GNNs\\ \cline{2-3}
                  &  \cite{guo2022learning, zhao2022understanding, Guo2023Generalizable, Wang2023Learning, Jiang2021graph, Multidimensional2023Liu,MishraGNN2024} & \begin{tabular}[c]{@{}c@{}} Heuristically design HetGNNs, which are hard to satisfy complicated permutation properties \end{tabular} \\ \cline{2-3}
                  &  Proposed & \begin{tabular}[c]{@{}c@{}} Systemically design HetGNNs according to the types of \\ vertices and edges, which can satisfy complicated permutation properties \end{tabular} \\
\hline\hline
\end{tabular}
\vspace{-1mm}
\end{table*}

\section{Simulation Results} \label{section: simulation}
In this section, we use the MCMU-power allocation and hybrid precoding problems to show the impact of exploiting permutation prior and topology prior on the sample complexity, space complexity, and size generalization ability.

The sample complexity of a DNN is the minimal number of samples required to train the DNN for achieving an expected performance. The space complexity is the number of trainable parameters in the corresponding fine-tuned DNNs. We do not compare the time complexity, which is not the focus of this paper.


We compare the VertexGNN and the EdgeGNN designed with the proposed methods (with legend {\em ``VertexGNN''} and  {\em ``EdgeGNN''}) with the following DNNs.
\begin{itemize}
\item {\emph {``PENN'':}}  \emph{This is the DNN with matched permutation property to a policy by introducing parameter sharing into FNN using the method in \cite{zaheer2017deep, SEquivariance,maron2019universality,hartford2018deep}, which does not exploit topology prior.}
\item {\emph {``${\text {VertexGNN}}_{\text {NP}}$'':}}  \emph{This is the VertexGNN using the updating equation in \eqref{eqn:update-vertex}, which can exploit topology prior but does not have permutation property.}
\item {\emph {``FNN'':}}  \emph{This is the FNN that does not exploit any prior}.
\end{itemize}

We do not provide the results of ${\text {EdgeGNN}}_{\text {NP}}$ that uses \eqref{eqn:update-edge} as updating equation, because its number of trainable parameters is extremely large that causes memory overflow.

\begin{figure}
\centering
\centerline{\includegraphics[width=1.6in]{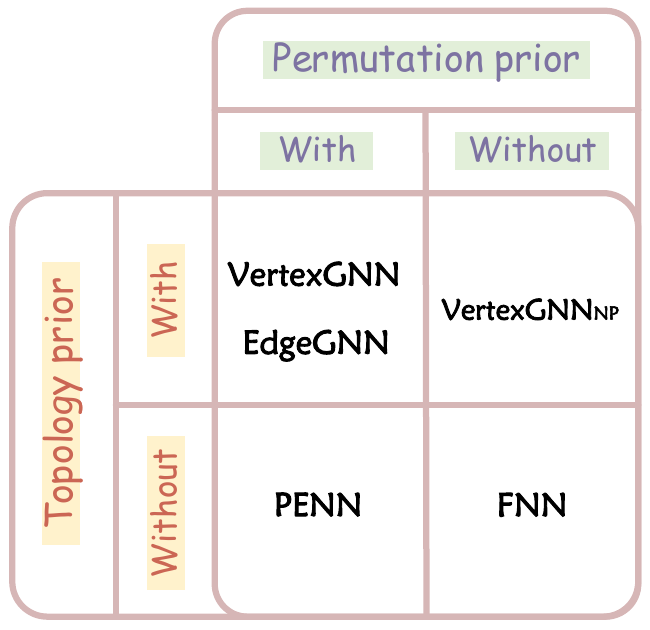}}
\caption{Relational priors exploited in each DNN}
\label{fig:rel-prior}
\end{figure}

In Fig. \ref{fig:rel-prior}, we summarize the exploited relational priors.

\subsection{Simulation Setup}

{\em \textbf{{ Example 1 (MCMU-Power Allocation):}}}
There are four cellular cells each with radius of 60 m. Each BS serves five randomly located UEs. The path loss is modeled as $32.4 + 31.9 \log_{10}(d) + 20\log_{10}(f_c)$, where $d$ is the distance between BS and UE in meters, $f_c$ is the carrier frequency and is set as 5 GHz. The shadowing is log-normal with standard deviation of 8 dB.
The small-scale channels follow Rayleigh fading. The maximal transmit power is 30 dBm and the noise power is -95 dBm.


{\em \textbf{{ Example 2 (Hybrid Precoding):}}} A BS with eight antennas and six RF chains serves two UEs. Each UE is either with a single antenna (i.e., MU-MISO system without ${\bf V}_{\rm RF}$) or with two antennas (i.e., MU-MIMO system).
The graph constructed for MU-MISO system contains three types of vertices (i.e., UE vertices, RF chain vertices and AN vertices) and three types of edges (the edges between every two types of vertices belong to a type).
The signal-to-noise ratio (SNR) is 10 dB.
The channel model and other setups are the same as in \cite{Multidimensional2023Liu}.


\subsection{Loss Functions and Fine-Tuned Hyper-parameters}
All DNNs are trained in an unsupervised manner \cite{S-unsuper}, where Adam algorithm and batch normalization are used.
For Example 1, the loss function is the negative SE in \eqref{eqn:problem-power}, and the activation function in (14) of \cite{guo2022learning} is used in the last layers of DNNs to satisfy the power constraint.
For Example 2, the loss function is negative SE in \eqref{eqn:obj3}. To satisfy the power and modulus constraints, we project ${{W}_{\tt RF}}_{ij}$ into $1/\sqrt{N_t} \cdot {{W}_{\tt RF}}_{ij}/|{{W}_{\tt RF}}_{ij}|$, ${{V}_{{\tt RF}, k}}_{i}$ into $1/\sqrt{N_r} \cdot {{V}_{{\tt RF}, k}}_{i}/|{{V}_{{\tt RF}, k}}_{i}|$, and ${\bf W}_{\tt BB}$ into $\sqrt{K} \cdot {\bf W}_{\tt BB}/|{\rm Tr}({\bf W}_{\tt RF}{\bf W}_{\tt BB}{\bf W}_{\tt BB}^{\sf H}{\bf W}_{\tt RF}^{\sf H})|$.

For {\em ``VertexGNN''} and  {\em ``EdgeGNN''}, the processors are respectively one-layer FNNs and linear processors, where the FNNs are used to avoid information loss \cite{PY}, the pooling and combining functions are respectively mean function and one-layer FNNs.
Other fine-tuned hyper-parameters of each DNN are summarized in Table \ref{table: para-dnns} (see next page).
\setlength{\tabcolsep}{8pt}
\renewcommand\arraystretch{1}
\begin{table*}[!htb]
\footnotesize
\centering
\caption{Fine-tuned hyper-parameters of the DNNs}\label{table: para-dnns}\vspace{-0.1mm}
\begin{tabular}{cc|c|c|c|c|c}
\hline
\multicolumn{2}{c|}{Hyper-parameters}                                       & VertexGNN & EdgeGNN & PENN & ${\text {VertexGNN}}_{\text {NP}}$ & FNN \\ \hline
\multicolumn{1}{c|}{\multirow{3}{*}{Power allocation}} & Learning rate       &    0.01       &  0.01       & 0.001 &  0.01 &  0.001             \\ \cline{2-7}
\multicolumn{1}{c|}{}                           & Number of neurons   &   [10]$\times$4        &  [10]$\times$4      &  [40]$\times$3   & [2]$\times$4  &   [1000]$\times$3            \\ \cline{2-7}
\multicolumn{1}{c|}{}                           & Activation function &  {\tt Leaky Relu}         & {\tt Tanh}       &  {\tt Relu}   &    {\tt Leaky Relu}  &   {\tt Relu}         \\ \hline
\multicolumn{1}{c|}{\multirow{3}{*}{\tabincell{c}{MU-MISO}}} & Learning rate       &  0.0002         &   0.0001        &      0.001 &    0.001   &  0.001        \\ \cline{2-7}
\multicolumn{1}{c|}{}                           & Number of neurons   &  [128]$\times$10      & [128]$\times$12        &  [1024]$\times$8    &     [32]$\times$12  & [1024]$\times$4     \\ \cline{2-7}
\multicolumn{1}{c|}{}                           & Activation function &  {\tt Softmax}         & {\tt Tanh}        &    {\tt Tanh}  &       {\tt Softmax}  &     {\tt Relu}    \\ \hline
\multicolumn{1}{c|}{\multirow{3}{*}{\tabincell{c}{MU-MIMO}}} & Learning rate       &  0.0002         &   0.0001        &      0.001 &    0.001  &  0.001         \\ \cline{2-7}
\multicolumn{1}{c|}{}                           & Number of neurons   &  [128]$\times$10      & [128]$\times$12         &  [2048]$\times$6    &     [128]$\times$6  & [3000]$\times$4     \\ \cline{2-7}
\multicolumn{1}{c|}{}                           & Activation function &  {\tt Softmax}         & {\tt Tanh}        &    {\tt Tanh}  &       {\tt Softmax} &     {\tt Relu}     \\ \hline
\end{tabular}
\vspace{-1mm}
\end{table*}

\setlength{\tabcolsep}{6pt}
\renewcommand\arraystretch{1.1}
\begin{table*}[!htb]
\centering
\footnotesize
\caption{({Sample complexity}, {space complexity})  of DNNs} \label{table: sample-space}
\begin{tabular}{c|c|c|c|c|c}
\hline
                 & VertexGNN & EdgeGNN & PENN & ${\text {VertexGNN}}_{\text {NP}}$ & FNN \\ \hline
Power allocation &    (4, 2.7k)     &     (4, 6.0k)   &  (10, 2.4k)    &  (20, 19.2k)  &   (500, 3420.0k) \\ \hline
MU-MISO &   (4000, 985.6k)      &  (800, 1179.6k)  & (2000, 589.8k)   &  (10000, 2073.6k) &  (32000, 8482.8k) \\ \hline
MU-MIMO &   (14000, 1966.1k)      & (6000, 3539.5k)   & (28000, 2291.8k)   & (30000, 20950.0k)  & (100000, 108480.0k) \\ \hline
\end{tabular}
\vspace{-4mm}
\end{table*}

\subsection{Results}
\subsubsection{Learning Performance} \label{section: learninig}
The learning performance is measured by the sum rate ratio, which is the ratio of the sum rate achieved by the learned policies to the sum rate achieved by numerical algorithms. In particular, we use weighted sum mean-square error minimization (WMMSE) algorithm \cite{shi2011iteratively} for Example 1, and GLRAM algorithm \cite{song2016Coordinated} for Example 2.

\begin{figure}[!htb]
\centering
\subfigure[Power allocation]{
\includegraphics[width=0.38\textwidth]{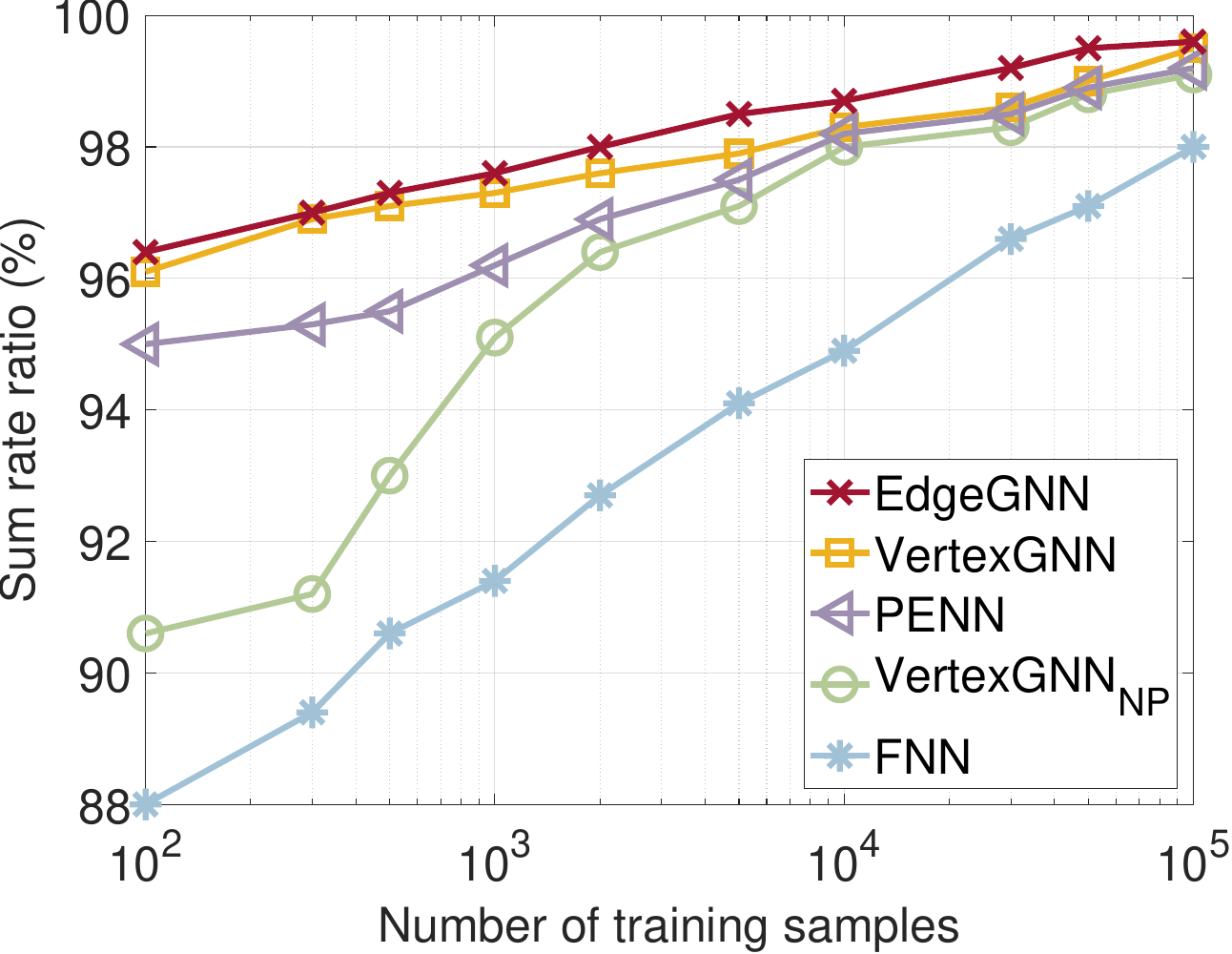}
\label{fig:learning-power}
} \\
\subfigure[Precoding, MU-MISO]{
\includegraphics[width=0.38\textwidth]{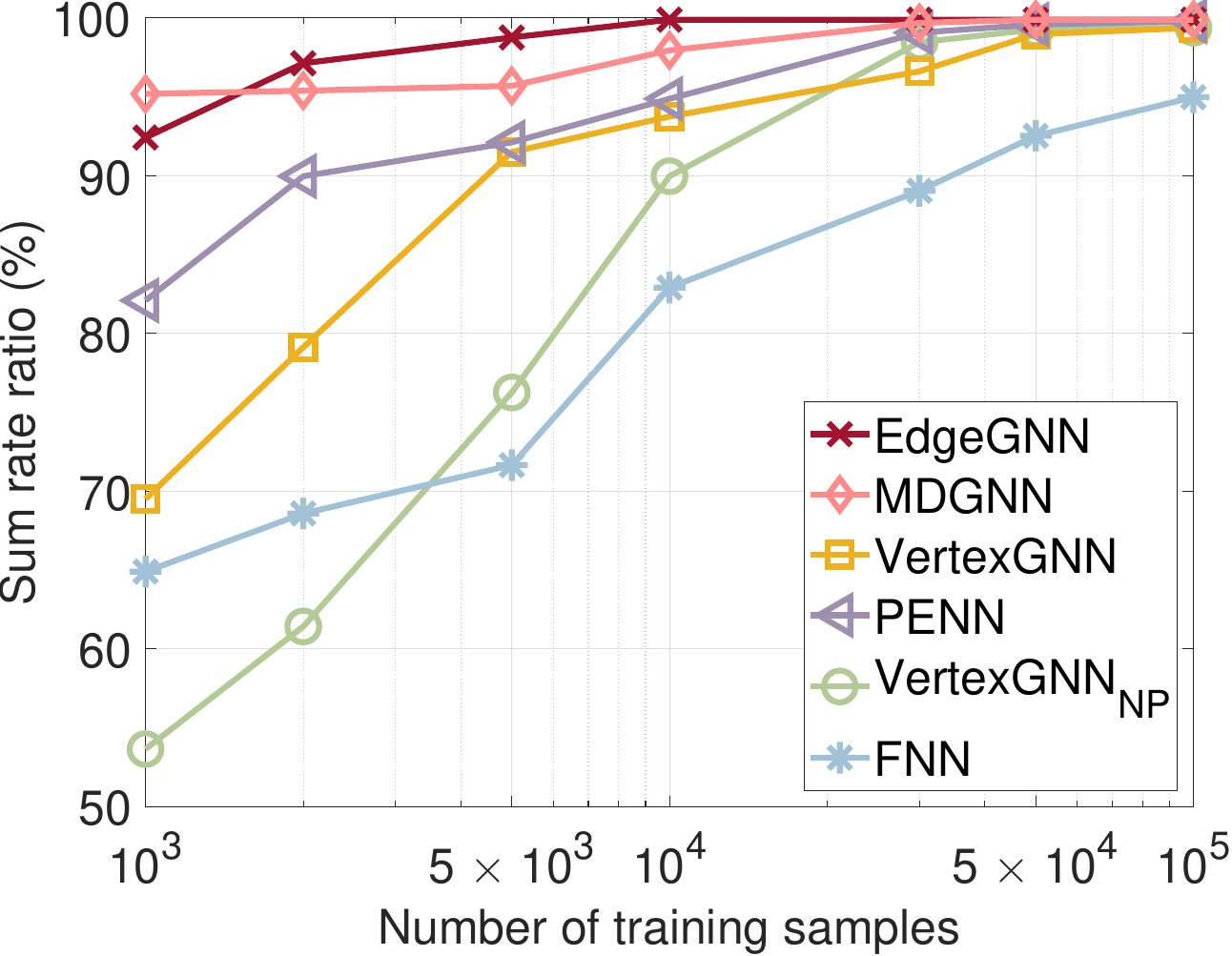}
\label{fig:learning-miso}
} \\
\subfigure[Precoding, MU-MIMO]{
\includegraphics[width=0.38\textwidth]{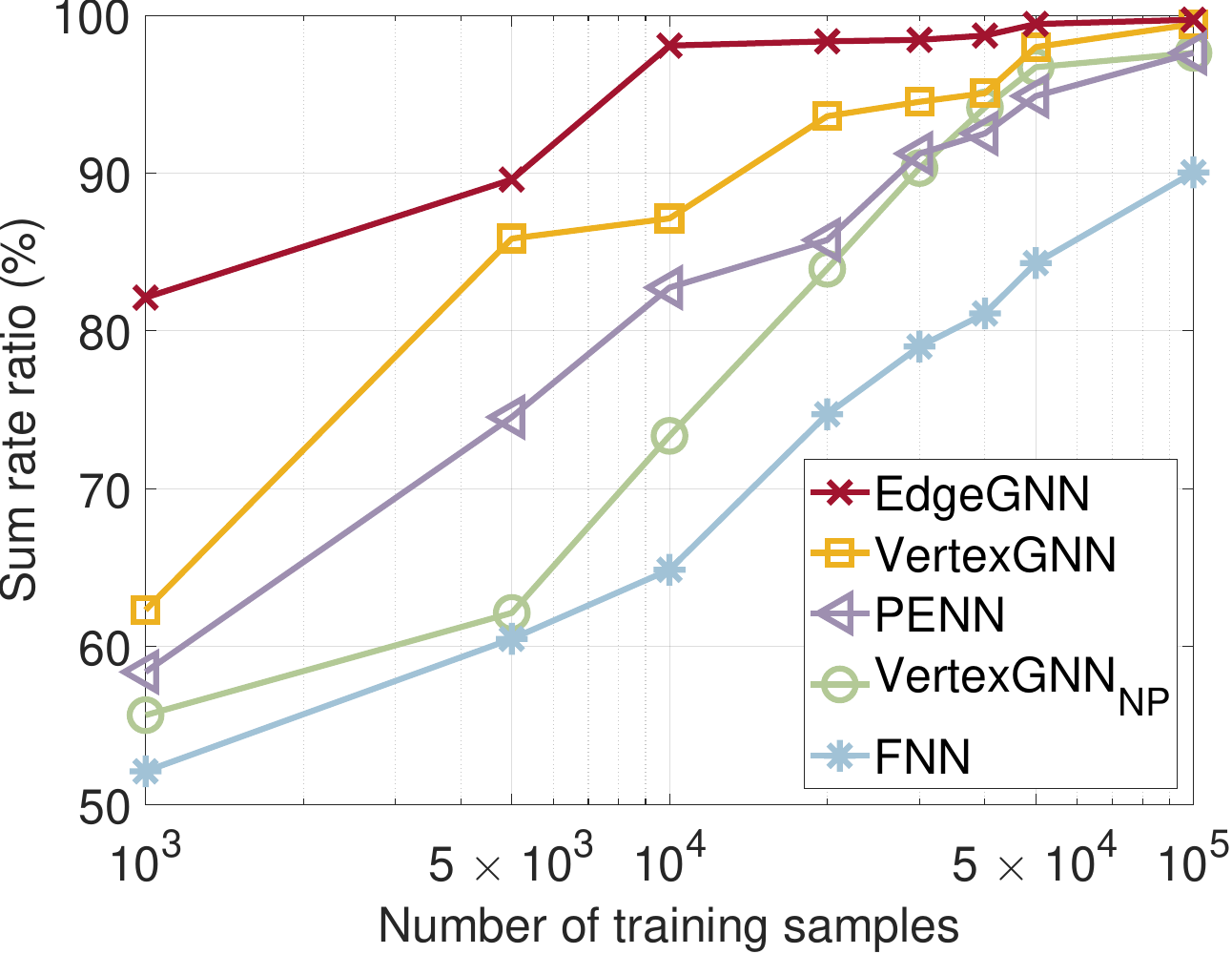}
\label{fig:learning-mimo}
}
\caption{Impact of relational priors on the learning performance of different policies with different numbers of training samples}
\label{fig:learnings}
\end{figure}

In Fig. \ref{fig:learnings}, we compare the learning performance of the DNNs versus the numbers of training samples. Due to the unaffordable training time of {\emph {``${\text {VertexGNN}}_{\text {NP}}$''}} and {\emph {``FNN''}}, we only provide the results with  up to $10^5$ samples.
We can see that {\emph {``PENN''}} and {\emph {``${\text {VertexGNN}}_{\text {NP}}$''}} require  fewer  training samples than {\emph {``FNN''}} to achieve the same sum rate ratio.
This indicates that exploiting either permutation prior or topology prior can reduce the sample complexity.
The {\emph {``PENN''}} needs fewer training samples than {\emph {``${\text {VertexGNN}}_{\text {NP}}$''}}. This indicates that exploiting permutation prior is more efficient than exploiting topology prior for the two problems. Both {\em ``VertexGNN''} and  {\em ``EdgeGNN''} outperform {\emph {``PENN''}} with fewer samples. This indicates that exploiting both priors can further reduce the sample complexity comparing with only harnessing one kind of prior.

Besides, we can see that {\em ``VertexGNN''} and  {\em ``EdgeGNN''} achieve almost the same performance when learning the power allocation policy, but {\em ``VertexGNN''} does not perform well when learning the precoding policy without sufficient samples.
This is because several FNNs are used as processors in {\em ``VertexGNN''}.
In Fig. \ref{fig:learning-miso}, we also provide the performance of the MDGNN in \cite{Multidimensional2023Liu} (with legend {\em ``MDGNN''}), which avoids the information loss by learning over a hyper-graph. It is shown that {\em ``EdgeGNN''} outperforms {\em ``MDGNN''}.
When learning over the graphs constructed for power allocation,  {\em ``MDGNN''} degenerates into {\em ``EdgeGNN''}, hence we do not provide the results of {\em ``MDGNN''} in Fig. \ref{fig:learning-power}.
Since {\em ``MDGNN''} incurs memory overflow when the numbers or types of vertices and edges are large, we also do not provide its performance for MU-MIMO in Fig. \ref{fig:learning-mimo}.

\subsubsection{Training Complexity}
In Table \ref{table: sample-space}, we provide the sample complexity  and the space complexity to achieve 90\% sum rate ratio. We can see that exploiting permutation prior is more efficient than exploiting topology prior, and exploiting one or both priors can reduce training complexity dramatically. {\em ``EdgeGNN''} is with higher space complexity than {\em ``VertexGNN''}, because there are more processors in {\em ``EdgeGNN''} than in {\em ``VertexGNN''}. In particular, the numbers of processors in {\em ``VertexGNN''} and {\em ``EdgeGNN''} are respectively 6 and 14 for power allocation, 6 and 12 for MU-MISO, and 8 and 14 for MU-MIMO. The space complexity of {\emph {``PENN''}} is lower than {\em ``VertexGNN''} and {\em ``EdgeGNN''}, because the HetGNNs are deeper than {\em ``PENN''} as shown in Table \ref{table: para-dnns}.

In Table \ref{table: topology}, we show the impact of exploiting topology prior in reducing both complexities considering that different policies are learned over the graphs with different sparsity. We use graph density to quantify the topology prior since it affects the number of zero sub-matrices in the updating equation of a GNN as exemplified in \eqref{eqn:updating-D2D}. The density is defined as ${\beta}=|\mathcal{E}|/|\mathcal{V}|$ \cite{kowalik2006approximation}, where $|\mathcal{E}|$ and $|\mathcal{V}|$ are respectively the number of edges and the number of vertices in a graph. A smaller value of $\beta$ indicates a sparser graph.
Considering that {\emph {``${\text {VertexGNN}}_{\text {NP}}$''}} only exploits topology prior but {\emph {``FNN''}} does not exploit any priors, we define $\eta_{\rm space}=\frac{{\rm space}_{\rm FNN}}{{\rm space}_{{\text {VertexGNN}}_{\text {NP}}}}$ to measure the impact of exploiting topology prior on reducing space complexity, where ${\rm space}_{\rm FNN}$ and ${\rm space}_{{\text {VertexGNN}}_{\text {NP}}}$ are respectively the space complexity of {\emph {``FNN''}} and {\emph {``${\text {VertexGNN}}_{\text {NP}}$''}}. $\eta_{\rm sample}$ is defined similarly.

From the table we can see that $\eta_{\rm space}$ and $\eta_{\rm sample}$ of power allocation are larger than those of precoding although the graph for power allocation is denser. This is because precoding policies are harder to learn than the power allocation policy \cite{zhao2022understanding}, hence wider and deeper {\emph {``${\text {VertexGNN}}_{\text {NP}}$''}} and {\emph {``FNN''}} are required to achieve the same performance as shown in Table \ref{table: para-dnns}.
This indicates that exploiting the topology prior for learning precoding policies is less beneficial in terms of reducing training complexity (i.e., with smaller values of $\eta_{\rm space}$ and $\eta_{\rm sample}$).
$\eta_{\rm space}$ and $\eta_{\rm sample}$ for MU-MISO and MU-MIMO systems are comparable, since the values of $\beta$ of the precoding graphs for the two systems are almost identical.

\vspace{2mm}
\renewcommand\arraystretch{1}
\begin{table}[!htb]
\centering
\small
\caption{Impact of exploiting topology prior} \label{table: topology}
\begin{tabular}{c|c|c|c}
 \hline
                 & ${\beta}$ & $\eta_{\rm space}$ & $\eta_{\rm sample}$  \\ \hline
Power allocation &    10.0     &     178.1   &  25.0    \\ \hline
Precoding in MU-MISO system &   4.75      &  4.1  & 3.2   \\ \hline
Precoding in MU-MIMO system &   4.8      & 5.2   & 3.3   \\ \hline
\end{tabular}
\end{table}

\subsubsection{Size Generalization}
\begin{figure}[!htb]
\centering
\begin{minipage}[c]{0.45\textwidth}
\centering
\includegraphics[scale=0.36]{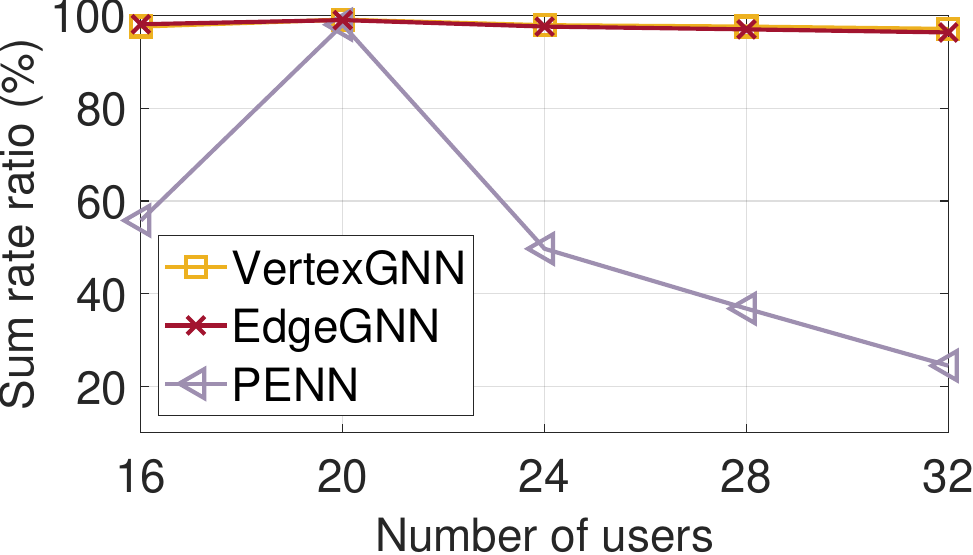}
\label{fig:gen-power}
\end{minipage}
\caption{Size generalization of DNNs, power allocation}
\label{fig:gen}
\end{figure}

In Fig. \ref{fig:gen}, we provide the size generalization performance of the HetGNNs and {\emph {``PENN''}} for learning the power allocation policy. Since the HetGNNs simply with desired permutation property cannot be generalized to the number of UEs for hybrid precoding as evaluated in existing literature, we do not re-produce the results.
Because {\emph {``${\text {VertexGNN}}_{\text {NP}}$''}} and {\emph {``FNN''}} do not satisfy any permutation property, they cannot be applied to different input sizes and hence their results are also not provided.

To evaluate the size generalizability, the DNNs are trained with samples generated in scenarios with 20 UEs and are tested in scenarios with $\{16, 24, 28, 32\}$ UEs. We can see that both HetGNNs can be well-generalized to the values of $K$ but {\emph {``PENN''}} cannot.
This indicates that exploiting both priors is necessary for enabling size generalizability.

\vspace{2.2mm}

\section{Conclusion}\label{conclu}
In this paper, we investigated how to model graphs and design HetGNNs to exploit the permutation properties of wireless policies. We found the essential role of edges and their type for HetGNNs on satisfying the permutation properties of the problems with nested sets or multiple related sets, and proposed a method to construct graphs.
We extended a method of introducing parameter sharing to FNN for satisfying given permutation property into the design of HetGNNs under an assumption. We proceeded to propose a method to design function sharing into HetGNNs without the assumption, which can satisfy desired permutation property by learning over the designed graph.
We obtained the following observations from the simulation results. (1) Exploiting permutation prior is more beneficial than topology prior for designing efficient DNNs in terms of low sample complexity and space complexity. Exploiting both priors is with lower training complexity than exploiting one of them. (2) Either only satisfying desired permutation property or only harnessing topology prior does not enable a DNN to be size generalizable.

\section*{APPENDIX Proof of Proposition \ref{pro1}}
\label{appb}
\setcounter{equation}{0}
\setcounter{subsection}{0}
\renewcommand{\theequation}{A.\arabic{equation}}
\renewcommand{\thesubsection}{A.\arabic{subsection}}

Denote $n_a$ as the number of vertices of the $a$-th type, where $a \in \mathcal{A}$ and $\mathcal{A}$ is the set of vertex types. The indices of all vertices and the indices of the vertices of the $a_1$-th type can be respectively expressed as,
\begin{equation}
\nonumber
\{ \underbrace{1,\cdots, n_1}_{\text{the first type}}, \underbrace{n_1+1,\cdots,n_1+n_2}_{\text{the second type}},\cdots, \sum_{a=1}^{|\mathcal{A}|}n_a\},
\end{equation}
and
\begin{equation}
\nonumber
{\mathcal{N}}_{a_1}=\{\sum_{i=1}^{a_1-1}{n_i}+1, \cdots, \sum_{i=1}^{a_1}{n_i}\}.
\end{equation}


We first consider that the vertices of the $a_1$-th type can be permuted independently.
When we permute the vertices of the $a_1$-th type, the indices of these vertices become $\{\pi(\sum_{i=1}^{a_1-1}{n_i}+1), \cdots, \pi(\sum_{i=1}^{a_1}{n_i})\}$.

Recall that ${\bf d'}_{m}^{(l)}$ and ${\bf d}_{m}^{(l)}$ are respectively the aggregation output and hidden representation of the $m$-th vertex in the $l$-th layer, and  $L$ is the number of layers of the GNN. We use $\widetilde{{\bf d'}}_{m}^{(l)}$ and $\widetilde{\bf d}_{m}^{(l)}$ to denote the aggregation output and hidden representation of the $m$-th vertex in the $l$-th layer when we permute the vertices of the $a_1$-th type.

We aim to prove that
\vspace{-1mm}
\begin{equation} \label{eqn:arb}
\widetilde{\bf d}_{\pi(m)}^{(l)}={\bf d}_{m}^{(l)}, \quad \forall m \in \mathcal{N}_{a_1}, \vspace{-1mm}
\end{equation}
are true for $\forall l$.
In the sequel, we use \emph{Mathematical Induction} method to prove \eqref{eqn:arb}.

When $l=0$, ${\bf d}_{m}^{(0)}$ is the feature of the $m$-th vertex. The features of the vertices in $\mathcal{N}_{a_1}$ are permuted in the same way as the permutation of indices, hence \eqref{eqn:arb} is true.

In what follows, we prove that \eqref{eqn:arb} is true when $l=l_0+1$ if \eqref{eqn:arb} is true when $l=l_0$.
According to \eqref{eqn:update-vertex}, if we do not permute the vertices of the $a_1$-th type, the hidden representation of the $m$-th vertex in the $(l_0+1)$-th layer can be obtained as
\begin{equation} \label{eqn:agg1}
\begin{aligned}
{\bf d}_{m}^{(l_0+1)}&={\rm CB}_{m}\!\left({\bf d}_{m}^{(l_0)}, {\bf d'}_{m}^{(l_0+1)};{\bm \theta}_{m}^{(l_0+1)}\right), \\
{\bf d'}_{m}^{(l_0+1)} \!\!\!& =\! {\rm PL}_{m}\!\!\left(\!q_{m, u}({\bf d}_{u}^{(l_0)}\!, {\bf e}_{m}^u;{\bm \phi}_{m, u}^{(l_0+1)}\!), u \!\in \! \mathcal{N}(m) \!\right)\!.
\end{aligned}
\end{equation}

${\rm CB}_{m}(\cdot,\cdot;{\bm \theta}_{m}^{(l_0+1)})$ in \eqref{eqn:agg1} maps ${\bf d}_{m}^{(l_0)}$ and ${\bf d'}_{m}^{(l_0+1)}$ into ${\bf d}_{m}^{(l_0+1)}$. If ${\rm CB}_{m}(\cdot,\cdot;{\bm \theta}_{m}^{(l_0+1)})$, ${\bf d}_{m}^{(l_0)}$ and ${\bf d'}_{m}^{(l_0+1)}$ remain unchanged when we permute the vertices of the $a_1$-th type, then ${\bf d}_{m}^{(l_0+1)}$ will be unchanged. ${\rm CB}_{m}(\cdot,\cdot;{\bm \theta}_{m}^{(l_0+1)})$ remains unchanged because of the condition that the combination function is the same for all the vertices with the same type. ${\bf d}_{m}^{(l_0)}$ remains unchanged because we assume that \eqref{eqn:arb} is true when $l=l_0$, i.e., $\widetilde{\bf d}_{\pi(m)}^{(l_0)}={\bf d}_{m}^{(l_0)}$. In the following, we prove that ${\bf d'}_{m}^{(l_0+1)}$ remains unchanged when we permute the vertices of the $a_1$-th type.

Considering that when we re-order a vertex (say the $m$-th vertex is re-ordered as the $\pi(m)$-th vertex and the two vertices belong to the $a_1$-th type), its neighboring vertices and the edges connecting them are re-ordered accordingly, i.e., ${\bf e}_{m}^{n}={\bf e}_{\pi(m)}^{n}, \forall n \in {\mathcal{\tilde  N}}(\pi(m))$, where ${\mathcal{\tilde N}}(\pi(m))$ is the set of neighboring vertices of the $\pi(m)$-th vertex after the re-ordering. Moreover, $q_{m, n}(\cdot;{\bm \phi}_{m, n}^{(l_0+1)})={q}_{\pi(m), n}(\cdot;{\bm \phi}_{\pi(m), n}^{(l_0+1)})$ due to Condition (C1) in Proposition \ref{pro1}. Then, we have
\vspace{-2mm}
\begin{equation}
\label{eqn:pro2}
\begin{aligned}
q_{m, n}({\bf d}_{n}^{(l_0)}, {\bf e}_{m}^{n};{\bm \phi}_{m, n}^{(l_0+1)})={q}_{\pi(m), n}({\bf d}_{n}^{(l_0)}, {\bf e}_{\pi(m)}^{n}; \\ {\bm \phi}_{\pi(m), n}^{(l_0+1)}),
\forall m \in \mathcal{N}_{a_1}, \forall n \in {\mathcal{\tilde  N}}(\pi(m)).
\end{aligned}
\end{equation}
\vspace{-3mm}

Under the condition that the pooling function is the same for all vertices with the same type, the pooling function of the $m$-th vertex and the $\pi(m)$-th vertex is the same.
According to \eqref{eqn:pro2}, and further considering that the pooling function satisfies the commutative law, we can obtain
\begin{equation}
\nonumber
\begin{aligned}
& {\rm PL}_{m}\left(q_{m, n}({\bf d}_{n}^{(l_0)}, {\bf e}_{m}^{n};{\bm \phi}_{m, n}^{(l_0+1)}), n \in \mathcal{N}(m)\right) = \\
& {\rm {PL}}_{\pi(m)}\left({q}_{\pi(m), n}({\bf d}_{n}^{(l_0)}, {\bf e}_{\pi(m)}^{n};{\bm \phi}_{\pi(m), n}^{(l_0+1)}), n \in {\mathcal{\tilde  N}}(\pi(m))\right).
\end{aligned}
\end{equation}
This indicates that ${\bf d'}_{m}^{(l_0+1)}$ remains unchanged when we permute the vertices of the $a_1$-th type.

From \eqref{eqn:agg1}, we know that ${\bf d}_{m}^{(l_0+1)}$ remains unchanged, i.e., \eqref{eqn:arb} is true for the $m$-th vertex when $l = l_0 +1$. Since the $m$-th vertex is arbitrarily chosen from $\mathcal{N}_{a_1}$, the proof holds for all the vertices in $\mathcal{N}_{a_1}$. Therefore, \eqref{eqn:arb} is true when $l = l_0 +1$.

By using {Mathematical Induction} method, we know that \eqref{eqn:arb} is true for $\forall l$.

Similarly, we can prove that \eqref{eqn:arb} is also true  when the vertices of the $a_1$-th type should be permuted with other types of vertices jointly, again by using the {Mathematical Induction} method.

This completes the proof for Proposition 1.

\bibliographystyle{IEEEtran}
\bibliography{jianyubib}

\end{document}